\RequirePackage{fix-cm}
\documentclass[smallcondensed]{svjour3}     
\smartqed  
\usepackage{multirow}
\usepackage{graphicx}
\usepackage{amsmath,amsfonts,amssymb}
\usepackage{wrapfig}
\usepackage{hyperref}
\usepackage{caption,subcaption}
\usepackage[ruled]{algorithm2e}
\usepackage{xcolor}
\newtheorem{thm}{Theorem}
\newtheorem{prop}{Proposition}
\newtheorem{defi}{Definition}
\newtheorem{lem}{Lemma}
\newtheorem{rem}{Remark}
\begin{document}

\title{Interpretable feature subset selection: A Shapley value based approach
}

\titlerunning{Interpretable feature subset selection: A Shapley value based approach}        

\author{Sandhya Tripathi        \and
        N. Hemachandra \and Prashant Trivedi 
}


\institute{ Sandhya Tripathi  \at 
	Department of Anesthesiology \\
		Washington Univeristy School of Medicine in St Louis, MO 63110\\
		\email{sandhyat@wustl.edu}
	 \and
        N. Hemachandra \and Prashant Trivedi  \at
                Industrial Engineering and Operations Research\\
        IIT Bombay,  Mumbai, India 400 076\\
        \email{\{nh,  trivedi.prashant15\}@iitb.ac.in}
}

\date{Received: date / Accepted: date}

\maketitle
 
\begin{abstract}
For feature selection and related problems, we introduce the notion of classification game, a cooperative game, with features as players and hinge loss based characteristic function and relate a feature's contribution to Shapley value based error apportioning (SVEA) of total training error.
Our major contribution is ($\star$) to show that for any dataset the threshold 0 on SVEA value identifies 
feature subset whose joint interactions for label prediction is significant or those features that span a subspace where the data is predominantly lying.
In addition, our scheme ($\star$) identifies the features on which Bayes classifier doesn't depend but any surrogate loss function based finite sample classifier does; this contributes to the excess $0$-$1$ risk of such a classifier, ($\star$) estimates unknown true hinge risk of a feature, and  ($\star$) relate the stability property of an allocation and negative valued SVEA by designing the analogue of core of classification game. Due to Shapley value's computationally expensive nature, we build on a known Monte Carlo based approximation algorithm that computes characteristic function (Linear Programs) only when needed.
We address the potential sample bias problem in feature selection by providing interval estimates for SVEA values obtained from multiple sub-samples. We illustrate all above aspects on various synthetic and real datasets and show that our scheme achieves better results than existing recursive feature elimination technique and ReliefF in most cases. Our theoretically grounded classification game in terms of well defined characteristic function offers interpretability (which we formalize in terms of final task) and explainability of our framework, including identification of important features.

\end{abstract}

\section{Introduction}
``What is the guarantee that a given model uses important and relevant features among the given features?'' This question has been the topic of research for decades in many learning areas, including supervised learning. To address this question in a binary classification task, we present a cooperative game-theoretic framework for feature subset selection. We introduce a classification game with features as players and hinge loss based characteristic function (in terms of linear programs, LPs).
As the training error of a classifier that does not use any features (players) is always non-zero, the challenge in defining a cost game is to deal with the requirement that characteristic function's value should be zero for the empty coalition.
We overcome this challenge by suitably defining a value game and apportioning the total training error of the hinge loss based linear classifiers using an affine transformation of the Shapley value of the value game. As Shapley value allocates the total training error to each  feature based on its proportional contribution (`paid as per your participation, no more, no less'), it is theoretically sound and has been famous as a cost allocation measure (\cite{fiestras2011cooperative,khare2015shapley,Kimms2016}) and in other areas as well (\cite{nisan2007algorithmic}). It also captures the interactions among features by the marginal contribution of a feature. Thus, it is a suitable choice for tasks like Feature Subset Selection (FSS).  
Further, Shapley value is a core selector in convex games and hence has desirable properties like stability in coalition formation.
Our \textbf{major contributions} are:
\begin{enumerate}
    \item Identification of features whose joint contribution to label prediction is significant, as given in Section \ref{subsec: neg_fss}. The feature subset can either be determined by a uniform threshold of 0 for all datasets on Shapley value based error apportioning (SVEA) value or by ranking the SVEA values for a user-given feature set size.
        \item We explicitly define the notion of an interpretable FSS scheme (Definition \ref{def: Interpre}) and evaluate a range of FSS scheme w.r.t. the proposed definition. We observe that our scheme satisfies all conditions required of an interpretable FSS scheme.
    \item Identification of features that span a subspace in which data lies with high probability and hence crucial for label prediction, as given in Section \ref{subsec: neg_dr}. Unlike many existing dimension reduction techniques, our scheme doesn't transform the feature space. {Working with a feature subspace, rather than a  transformed space,  is desirable for model interpretability and explainability}.
    \item In Section \ref{subsec: eta_error_decomp} and \ref{subsec: converg_SV_error}, we provide insights on the contribution of features whose SVEA value is positive, to the excess $0$-$1$ risk of a surrogate loss based classifier. We also provide an estimate of the unknown true hinge risk of each feature.
    \item To capture stable apportioning of training error in classification tasks, we introduce the set $C_E(m)$ that is analogous to the core and hence useful for feature selection. We also study convex classification games as Shapley values are core selectors for such games and relate them to FSS (Section \ref{subsec: stable_app}).

\end{enumerate}

In addition to the above-listed contributions, we present a sampling-based approximation algorithm built on \cite{castro2009polynomial} that does not require computing characteristic function (LP) for $2^{n}$ subset of features all at once; instead, compute it only when a particular subset of features is sampled. Also, in Section \ref{subsec: sample_bias_robust_tech}, we attempt to address the sample bias issue by averaging over the Shapley value based error apportioning across multiple sub-samples and provide $t$-distribution based confidence intervals. We also considered another variant where the linear classifier based training error is regularized and computationally observed that the feature subset selected is the same as that of the unregularized model (details in Supplementary Material (SM) D.3).

Our idea that thresholding the modified Shapley value of classification game at 0 identifies the features with substantial joint contribution to the prediction has following motivation.
Suppose among a group of players (features), one player has sufficient resources so that it has the power to work (classify) alone. Let us call it a dominant player. Now, if the other players (features) ask this dominant player to join their coalition (to form a classifier), then it asks them for a payoff. Since, the quantity to be divided is an error (cost), for such dominant players, the payoff is in the form of modified Shapley value being negative. We demonstrate this phenomenon in Pima Diabetes dataset where, knowing the blood sugar level (feature) is sufficient to decide whether the patient has diabetes or not. 

An innate understanding of how our SVEA  scheme possesses explainability and interpretability (formally in Definition \ref{def: Interpre}) is as follows. 
Explainability in our scheme refers to its ability to provide a reason for selecting a feature as important using its SVEA value; a feature with negative SVEA lowers the total training error. An important feature subset constituting such features makes FSS (using the SVEA scheme) explainable. 
{Interpretability in the context of the SVEA scheme includes accounting for possible interactions among features using Shapley value, using the training error similar to the one used in final classification task and mapping SVEA (importance) value of a particular feature to an apportioning of training error by the Shapley value of  the well defined classification game.}
The SVEA values can either be negative or positive; features with negative value can be interpreted as the dominant ones (more details in Section \ref{subsec: neg_fss}). \\

\subsection{Related work}  \label{subsec: related_work}
In this section, first, we provide some work on feature subset selection. Recursive feature elimination by \cite{kohavi1997wrappers} and ReliefF by \cite{kononenko1997ReliefF} are the most popular wrapper and filter methods for feature subset selection. Recently, \cite{song2013fastFSS} present a graph-theoretic clustering-based FSS scheme that first clusters the features and then chose a representative from each cluster to get the final important feature set. An interesting idea of instance dependent FSS for a general task (classification or regression), is presented in \cite{cancela2019scalableInstanceFSS}, where the authors compute saliency for each feature by identifying a task and loss dependent gain function.

Cooperative game theory provides a compelling framework for understanding the influence of a single feature or their interactions on the label/class.  With this motivation, we give a brief overview of how cooperative game theory has been applied to solve various sub-problems arising in classification. To avoid any misunderstandings, we would like first to present how our work is different (in purpose and approach) from some existing work, which uses similar ideas, along with other contributions as well. 

Our explicit game formulation in terms of training error (natural approach of ERM) is novel; this further leads to {\it understanding of non-important features' contribution to excess 0-1-risk} and provides an { \it estimate of true hinge risk of each feature}. If the machine learning pipeline consists of 3 components, pre-processing steps like feature selection, actual learning of the hypothesis (classifier), and implications (explainability, transparency), then \cite{kononenko2010efficient} and \cite{oakland16_Datta} deal with the last component and \cite{cohen2005feature} deals with the first component of the pipeline. Explainability, transparency, and interpretability related work use the contribution of a feature (via Shapley value) to understand the reasons for a given classifier making a particular decision; in contrast, our objective is to use Shapley value to identify a subset of features important for predictions or the feature dimensions spanning the dataspace, and hence, belongs to the first component of the pipeline. \cite{cohen2005feature} uses Shapley value on top of the iterative wrapper technique, whereas we only use Shapley value (no iterations). Also, the former technique requires user given threshold on the contribution value whereas for us the threshold of 0, to decide feature subset is not user-given or tuned for but instead decided by the SVEA (indirectly using  $tr\_er(\emptyset, m)$ while apportioning). Explicitly, our work can be related to broadly following five research areas.
\\
\textbf{Cooperative game theory in classification and related tasks:} \cite{torkaman2011approach} used Shapley value to get the weights for a classifier. \cite{fragnelli2008game} used cooperative game theory where the classifier used is predefined, and Shapley value is directly proportional to the power of a feature (gene).
\\
\textbf{Cooperative game theory in feature selection:} Based on the search strategy used, \cite{chandrashekar2014survey} classifies the feature selection techniques into three categories viz., filter techniques, wrapper techniques, and embedded methods. \cite{cohen2007feature} proposed a contribution selection algorithm that uses Shapley value to improve upon wrapper techniques like backward elimination and forward selection. \cite{sun2012ShapleyValue} and \cite{sun2012BanzafIndex} used Shapley value and Banzhaf index respectively to compute the importance of features which is further used with the filter methods based on information-theoretic ranking criteria. All the methods mentioned above use cooperative game theory mainly to give additional information to either a wrapper or filter method. 
{However, cooperative game theory is central to our scheme as it uses an affine transformation of Shapley value of the classification game which further provides interpretability and explainability to the selected feature subset.}
\\
\textbf{Cooperative game theory for explaining a prediction:} \cite{kononenko2010efficient} provide a Shapley value based explainability scheme and use feature contribution for explaining prediction for a given data point; here feature's contribution can change when a different data point is used. 
An axiomatic approach based on cooperative game theory to define an influence measure is available in \cite{datta2015influence}.  Instead of training classifiers, the influence measure is used to decide which features influence the decision of an unknown classifier. \cite{lundberg2017unified} define a class of additive feature attribution methods and use game theory results to explain a model's prediction. \cite{sundararajan2017axiomatic} presented an axiomatic attribution approach for deep neural networks. Unlike \cite{kononenko2010efficient} and other explainability methods, we are interested in apportioning the total training error using Shapley value allocation. Based on this apportioning, we want to differentiate the features as {\em essential} and {\em inessential}. Due to this, the training error functions and value functions in our approach are entirely different from the models in the current literature; we also provide various other interesting interpretations of our model.  
\\
\textbf{Cooperative game theory for data valuation:} Quantifying the value of data in algorithmic predictions and decisions has become a fundamental challenge. \cite{jia2019towardsvaluationAISTATS} study the problem of data valuation by utilizing the Shapley value and propose two Shapley value estimation algorithms that exploit the structure of the utility function of the game. \cite{ghorbani2019dataShapleyICML} also proposes Monte Carlo and gradient-based methods to efficiently estimate data Shapley values in practical settings where complex learning algorithms, including neural networks, are trained on large datasets. In addition, the authors claim that their methods can identify outliers and corrupted data and provide suggestions on how to acquire future data to improve the predictor. For both the above works, the player set for the underlying game is the set of total data points available. This is in contrast to our scheme, where we use the feature set as the player set. \cite{agarwal2019marketplaceEC} proposes a mathematical model of a system design for a data marketplace. They use Shapley value to divide the generated revenue ``fairly'' among the training features, so sellers get paid for their marginal contribution.

The difference between our work of using CGT for FSS (\textit{before training}) and existing work using it for explaining predictions (\textit{after training}), has been also clarified and highlighted by \cite{sundararajan2019shapley}, where they treat feature importance across all the training data and attribution (explaining model prediction) as two separate problems. CGT based data valuation work focuses on selection of the most relevant data points to apportion the overall profit among various contributors by considering data points as players, unlike our work, where we model features as players.
We would like to emphasize that our scheme is not just a global version of Shap \cite{lundberg2017unified} as the later scheme averages over the Shap values for every feature across data points to get a summary importance. Instead, we have an explicit game formulation (whose relevance has been already pointed out by \cite{merrick2020explanation} in a different setup of explaining predictions) whose Shapley values are used as feature contribution to the training error. Hence, our definition is novel and takes natural approach of Empirical Risk Minimization (ERM).

\textbf{Interpretability in feature subset selection:} 
Feature subset selection being an integral part of any learning model needs interpretable and explainable methods too. A visual explanation and interpretation approach for dimension reduction is presented by \cite{cavallo2018visual}. Mutual information based feature selection method that uses the unique relevant information and show its importance in health data is given in \cite{liu2018suri}. Local information based interpretable feature subset selection is also studied by \cite{yooninterpretable}. Another more recent approach called Informative Variable Identifier (IVI) in FSS by ensemble category is proposed by \cite{munoz2020informative} where they also provide levels of interpretability in FSS. However, their scheme relies on statistical properties of feature distribution to incorporate feature interactions. We provide a single definition for an FSS algorithm to be interpretable in Definition \ref{def: Interpre}. FSS methods that mitigate the bias amplification in linear models have been proposed by \cite{leino2019feature} wherein the authors have presented two new feature selection algorithms for mitigating bias amplification in linear models, and show how they can be adapted to convolutional neural networks efficiently. The influence function is used to remove the features which have bias towards the prediction. However, our focus is on the feature subset selection. We have also addressed the issue of sample biasedness, but that is different from the biasness of the features towards the prediction.

\subsection{Preliminaries}\label{subsec: preliminary}
In this section, we introduce some classification (\cite{Mohri}, \cite{steinwart2008support}) and cooperative game (\cite{Narahari},\cite{peleg2007introduction}) terminology and concepts to provide a better understanding of the connection which we will be studying in rest of the paper. 

\textbf{Classification setup:} Let $\mathcal{X}$ be the feature space and $\mathcal{Y}$ be the label set. Let $\mathcal{D}$ be the joint distribution over $\mathbf{X}\times Y$ with $\mathbf{X} \in \mathcal{X} \subseteq \mathbb{R}^{n}$ and $Y\in \mathcal{Y} = \{-1,1\}.$ Let the in-class probability and class marginal on $\mathcal{D}$ be denoted by $\eta(\mathbf{x}):=P(Y=1|\mathbf{x})$ and $\pi := P(Y=1)$ respectively. Let the decision function be $f:\mathbf{X}\mapsto \mathbb{R}$ and hypothesis class of all measurable functions be $\mathcal{H}$. {We consider linear hypothesis class $\mathcal{H}_{lin}= \{ (\mathbf{w},b), \mathbf{w}\in \mathbb{R}^n, b \in \mathbf{R}\}$ for all real dataset based experiments in this paper. This is because for non-linear models the basic assumption for Shapley value is violated; more details are provided in Section \ref{subsec: kernel_reg_ntworking_exp}.} We have an i.i.d. sample of size $m$ from distribution $\mathcal{D}$, viz., $ D = \{(\mathbf{x}_i,y_i)\}_{i = 1}^{m}$ where $\mathbf{x}_i = (x_{i1}, x_{i2},\ldots, x_{in})$ is the value of the feature and $y_i \in \{-1,1\}$ is the label for $i^{th}$ data point. We use hinge loss based Empirical Risk Minimization (ERM) setup because in addition to many desirable properties such as classification calibration and large margin it imparts to classifiers, it leads to an LP  which can be solved in polynomial time.

\textbf{Cooperative game theory} (\cite{Narahari,peleg2007introduction}): The Transferable Utility (TU) cooperative game is a pair $(N,v)$ where $N=\{1,\ldots,n\}$ is a set of players and $v: 2^N \mapsto \mathbb{R}$ is the characteristic function, with $v(\emptyset)= 0.$
Shapley value (\cite{Narahari,peleg2007introduction}) is a unique, symmetric, and strongly monotonic solution concept defined as a mapping $\phi : \mathbb{R}^{2^n-1} \mapsto \mathbb{R}^n$  given below:
\begin{equation*} \label{eq: Shapley value_cost}
\phi_j(v) = \sum_{S\subseteq N\backslash \{j\}} \frac{|S|!(n-|S|-1)!}{n!}[v(S\cup \{j\}) - v(S)],  ~~\forall j \in N, ~~\forall ~v \in 
\mathbb{R}^{2^n-1}.
\end{equation*} 
Detailed interpretation of Shapley value axioms from classification perspective is available in SM B.
\section{Training error based classification game $(N,v(\cdot, m))$} \label{sec:training_error_game}
In this section, we define the training error incurred by using a subset of features for classification based on the standard ERM setting. Next, we describe a value game and relate it to the training error via a one-to-one mapping. 

\subsection{Training error function}\label{training_error_fn}
Given the dataset/sample $D = \{(\mathbf{x}_i,y_i)\}_{i=1}^m$ with features $N = \{1,\ldots,n\}$, we consider a training error function, $tr\_er(S,m)$ associated with all possible subsets $S\subseteq N$ when sample size is $m$. 
We define $tr\_er(\emptyset, m)$ as hinge loss based training error of an intercept only classifier and denote it by $\tilde{c}(m):= tr\_er(\emptyset, m)$.

\begin{equation}
	\begin{aligned}
		\label{empty_lp}
		&tr\_er(\emptyset, m) =\min\limits_{b, \{\xi_i\}_{i=1}^m} \frac{1}{m}\sum\limits_{i=1}^{m}\xi_i \\
		& \text{s.t. } y_i b \geq 1 - \xi_i, ~~ \forall i = 1,\ldots, m\\
		& \xi_i \geq 0, ~~ \forall i = 1,\ldots,m.
	\end{aligned}
\end{equation}
 Similarly, we define the training error, $tr\_er(S,m)$ for any nonempty subset $S= \{j_1,j_2,\ldots,j_r\}$ of size $r$ with $r$ distinct elements/features. 
This would be minimal hinge loss of the classifier $  (w^{\ast}_{j_1},\ldots,w^{\ast}_{j_r},b^{\ast}_r)$ obtained from the dataset projected to $r$-dimensional subspace, i.e., dataset having feature values $\{x_{ij_1},\ldots,x_{ij_r}\}_{i=1}^m$ and label $\{y_i\}_{i=1}^m$. 
\begin{equation}
	\begin{aligned}
		\label{S_cost_lp}
		&tr\_er(S,m) = \min\limits_{w_{j_1},\ldots,w_{j_r},b_r,\{\xi_i\}_{i=1}^m} \frac{1}{m}\sum\limits_{i=1}^{m}\xi_i \\
		& \text{s.t. } y_i \left(\sum\limits_{j \in S}w_{j}x_{ij} + b_{r}\right)  \geq 1 - \xi_i, ~ \forall i \in [m]\\
		& \xi_i \geq 0, ~~ \forall i = 1,\ldots,m.
	\end{aligned}
\end{equation}

When $S=N$, we have $tr\_er(N,m)$ that is the minimal hinge loss based empirical risk of the classifier $(\mathbf{w}_N^{\ast},b^{\ast}_N)$ when the given dataset is $n$ dimensional, i.e. all $n$ feature values from the sample $D$ are used.
Note that the variables used in each ERM are local to that optimization problem only. 

As conventional cooperative games assume $v(\emptyset)=0$, training error function $tr\_er(\cdot,m)$ with $tr\_er(\emptyset,m) \neq0$ cannot be a valid characteristic function.
To circumvent this problem, we define a payoff/value game with characteristic function $v(S,m)$\footnote{Characteristic function as defined here depends on sample size $m$, so we use $m$ as an argument in $v(\cdot, m)$.}  given below:
\begin{equation}\label{value_to_cost}
    v(S,m) = tr\_er(\emptyset,m) - tr\_er(S,m), ~ \forall~ S \subseteq N.
\end{equation}
$v(S,m)$ represents the marginal improvement in the training error obtained due to the presence of the features in $S$. As, $v(\emptyset,m)=0$, it is a valid characteristic function also. This characteristic function along with the feature set $N$ defines a TU \textbf{classification game} $(N,v(\cdot,m))$. Further, the characteristic function $v(S,m)$ is monotonic w.r.t the coalitions, which is an important property from the perspective of allocation. This property is formalized in Proposition \ref{prop: game monotonic} with proof being available in SM  B.1.

\begin{prop} 
\label{prop: game monotonic}
If $(N,v(\cdot,m))$ is a classification game, then the characteristic function $v(\cdot,m)$ is monotonic, i.e.,
\begin{equation}\label{v_structure_1}
\forall ~~ S \subseteq T \subseteq N, ~~ v(S,m) \leq v(T,m).
\end{equation}
\end{prop}


\subsection{Training error allocation using Shapley value} \label{subsec: train_error_shapley}
As Shapley value solution concept has the idea of allocation based on a feature's marginal contribution (no more, no less), it emerges as a suitable candidate for apportioning of $v(N,m)$ among the features in a classification game. 
The Shapley value of classification game is given below:
\begin{equation} \label{eq: Shapley value}
\phi_j(N,v(\cdot,m)) =  \sum\limits_{S\subseteq N\backslash \{j\}} \frac{|S|!(n-|S|-1)!}{n!}[v(S\cup \{j\},m) - v(S,m)], ~\forall j \in N.
\end{equation}
Using this Shapley value, Theorem \ref{thm: appor} provides an equitable training error allocation among features.
We refer to it as  {\bf Shapley value based error apportioning  (SVEA)} denoted by $e_j(tr\_er(N,m)), ~\forall ~ j \in N$; as we see below, it is an affine transformation of Shapley value for feature $j\in N$. A proof of Theorem \ref{thm: appor} is available in SM A.1.
\begin{thm} \label{thm: appor}
There exists a Shapley value based error apportioning, $e: \mathbb{R}^{2^n-1} \rightarrow \mathbb{R}^n$  of the total training error among the features as given in the expression below:
\begin{equation}\label{eq: apport_total_train_error}
e_j(tr\_er(N,m)) = \frac{\tilde{c}(m)}{n} - \phi_j(N,v(\cdot,m)), ~~ \forall j ~ \in N.
\end{equation}
\end{thm}
For notational convenience, hereafter, we will denote the allocation of training error to feature $j$, by $e_j(m)$ and Shapley value of feature $j$ by $\phi_j(m)$ when the sample size is $m.$

In general, the problem of computing Shapley value is known to be NP-hard \cite{faigle1992shapley}. Also, it has high space complexity due to the space requirement of storing $n!$ permutations or $2^n -1$ characteristic functions. To bypass this issue, we adapt the approximation algorithm given by \cite{castro2009polynomial} for computing the Shapley value of features in the classification game $(N,v(\cdot,m))$. The advantage of using this algorithm is that characteristic function is calculated for a coalition as and when required in the marginal contribution sum. {Note that the computation of required $tr\_er(S,m)$ for a coalition $S$ is scalable as it is by an LP.} Algorithm and related details are available in SM C. 
In Section \ref{subsec: real_neg_SV_exp_feature}, we use this approximation for datasets with $n\geq 10$. {To evaluate the quality of Shapley value estimates, we compute their difference from the true Shapley value for datasets with $n<10$ and observed that use of 100 Monte Carlo (MC) samples lead to a min 0.5 \% and max 10\% error over 10 trials (different train and test partitioning) across all datasets. If the MC samples are increased to 1000, this error comes down to a max 4 percent. Since only the sign of the apportioning via approx Shapley value matters, use of 100 MC sample is sufficient as the sign is not affected (tested empirically for more than 100 samples too).}

\subsection{Properties of the classification game $(N,v(\cdot,m))$}
As shown in Proposition \ref{prop: game monotonic} the characteristic function, $v(S, m)$ is monotone as a function of feature set $S$. Next, a rational player (feature) joins a coalition only if it gets better than what it would get individually (by not forming a coalition), never less (Individual Rationality (IR)).  Also, there shouldn't be any surplus from the total value after allocation among the players (Collective rationality). 
As we are ultimately interested in SVEA, $e_j(m), ~j\in N$, we show that it is IR in Theorem \ref{thm: apportion_train IR} (Proof is available in SM A.2).
\begin{thm}
\label{thm: apportion_train IR}
Shapley value based error apportioning $\{e_j(m)\}_{j\in N}$ satisfies individual rationality, i.e., 
\begin{equation}\label{eq:IR_total train error}
    e_j(m) \leq tr\_er(\{j\}), ~\forall ~ j \in N.
\end{equation}
\end{thm}
This implies that the allocation of total training error to a feature is less than the training error had it been used alone, which is desirable and reasonable to expect. 
We observed on UCI datasets that various game theoretic properties for classification game don't hold universally and are dataset dependent. A detailed description of these properties along with counter examples is available in SM B.

\section{Insights from Shapley Value based Error Apportioning (SVEA) approach} \label{sec: SVEA_implications}
As we are interested in apportioning of $tr\_er(N,m)$ among features, it is possible that for some feature $j\in N$, $e_j(m)<0$. The intuition is as follows: suppose a player (feature) is so dominant that it can work (classify) alone. Now, if the other players (features) ask this dominant player to join their coalition (to form a classifier), then it asks them for a payoff. Since, the quantity to be divided is an error (cost), for such dominant players, the payoff is in the form of SVEA being negative. 
We formally present this idea in Proposition \ref{prop: 2d_SV_neg} for the two player case whose proof is available in SM A.3.

\begin{prop} \label{prop: 2d_SV_neg}
Consider a 2-feature classification game $(N,v(\cdot,m))$ with training error function $tr\_er(\{1\},m) = g >0$, $tr\_er(\{2\},m)= G >0$, $tr\_er(\{1,2\},m) = g^{\prime} \leq \min\{g,G\}$. If $\frac{G}{2} > g$, then SVEA of $tr\_er(\{1,2\},m)$ is such that $e_1(m) \leq 0$ and $e_2(m) \geq 0.$
\end{prop}
If we generalize the notion of Proposition \ref{prop: 2d_SV_neg} then, apportioning $\{e_j(m)\}_{j\in N}$ of  $tr\_er(N,m)$ can provide us with various insights. Based on the above arguments, we study the role of those features for which SVEA is negative, in FSS, in dimension reduction, and in excess $0$-$1$ risk decomposition of a finite sample-based classifier. Also, SVEA values can be interpreted as estimates of true unknown hinge risk of a feature.
Based on the above arguments, in this work, we study the role of those features for which SVEA is negative in feature subset selection. In particular, we show SVEA based decision of whether a feature is to be selected or not is easily interpretable. To formalize the notion of interpretabilty in FSS, we define it as follows:
\begin{defi} \label{def: Interpre}
	A scheme for FSS is said to be interpretable if it satisfies following conditions:
	\begin{enumerate}
		\item[(P1)] \textbf{Transparency in the process:} The process of finding the feature importance should be transparent and it should be clear how the feature interactions are being accounted for.
		\item[(P2)] \textbf{Relation to final task:} The feature importance computation should be based on a criterion which takes into account their role in final task (classification).
		\item[(P3)] \textbf{Justifiable importance values:}
		Feature importance values should have a meaning/justification in the context of the final task in addition to being just called feature contributions.
	\end{enumerate}
\end{defi}

\begin{figure}[h!]
	\centering
	\includegraphics[width=0.8\textwidth]{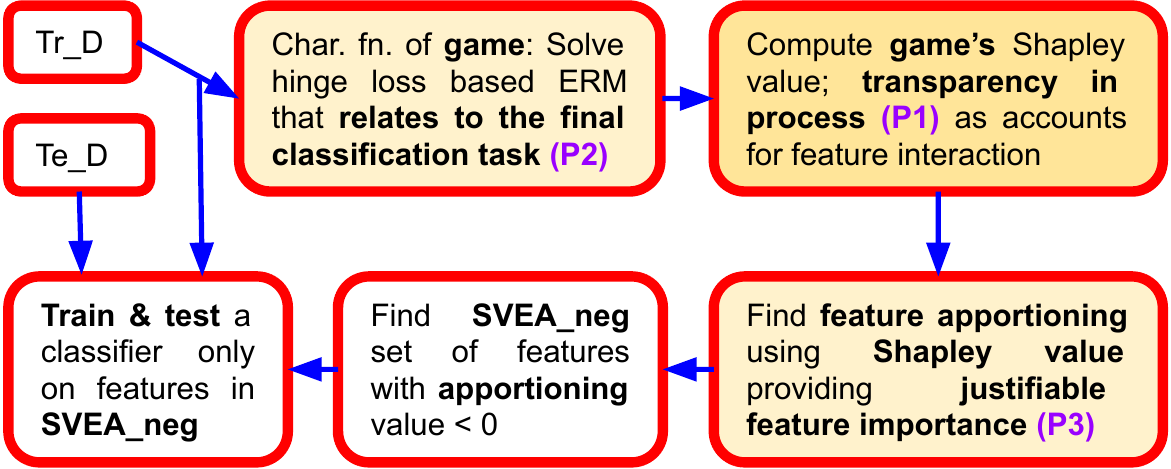}
	\caption{Flow chart describing interpretable FSS scheme SVEA with highlighted properties, P1, P2, and P3 defined in Def. \ref{def: Interpre}. Here, Tr\_D and Te\_D denote train and test dataset.}
	\label{fig: overall_Scheme}
\end{figure}

SVEA is interpretable as it satisfies all above conditions. It accounts for all possible interactions among features using Shapley value (P1), uses the training error similar to the one used in final classification task (P2) and SVEA (importance) value of a particular feature correspond to an apportioning of training error by the Shapley value of  the \emph{well defined} classification game (P3). Figure \ref{fig: overall_Scheme} depicts the steps of SVEA scheme in which the above properties are satisfied. Next, we summarize which FSS methods satisfy the different conditions from Definition \ref{def: Interpre} in Table \ref{tab: FSS_inter_methods}.

\emph{Remark:} The above definition of interpretablilty is fairly generic, as it can be adopted for other schemes (other than FSS) in a broader task (other than classifier design).


\begin{table}[h!]
	\centering
	\begin{tabular}{|c|c|c|c|}
		\hline
		\textbf{Methods/Conditions} & \textbf{P1} & \textbf{P2} & \textbf{P3} \\ \hline
		\textbf{ReliefF \cite{kononenko1997ReliefF}} & \checkmark   & \checkmark & $\times$ \\ \hline
		\textbf{RFECV \cite{kohavi1997wrappers}} & \checkmark & \checkmark & $\times$ \\ \hline
		\textbf{BanzhafI based \cite{sun2012BanzafIndex}} & \checkmark & \checkmark & $\times$ \\ \hline
		\textbf{ShapleyV based \cite{sun2012ShapleyValue}} & \checkmark & \checkmark & $\times$ \\ \hline
		\textbf{LFS \cite{yooninterpretable}} & \checkmark & \checkmark & $\times$ \\ \hline
		\textbf{IVI \cite{munoz2020informative}} & \checkmark  & \checkmark & $\times$ \\ \hline
		\textbf{SVEA based FSS \cite{munoz2020informative}} & \checkmark & \checkmark & \checkmark \\ \hline
	\end{tabular}
	\caption{Table summarizing FSS methods w.r.t. interpretability from Definition \ref{def: Interpre}.}
	\label{tab: FSS_inter_methods}
\end{table}

{Before proceeding further, we would like to note the difference between FSS and dimension reduction considered in this paper. FSS is a special case of dimension reduction, and to perform FSS, we use the SVEA scheme where feature subset to be used for the final classifier is identified based on the sign of SVEA value of a feature (Section \ref{subsec: neg_fss}).
However, there can be scenarios when some feature dimensions do not play a role in \textit{spanning the dataspace} as explained in Section \ref{subsec: neg_dr}; this is irrespective of learning task at hand, as seen in Fig \ref{fig:dim_red_2d} with the 2 dimensional data space being effectively spanned by 1 dimension, i.e., $x_1$.}

\subsection{Negative valued SVEA and FSS} \label{subsec: neg_fss}
We observed that the features for which SVEA is negative (set $SVEA_{neg}$) are the ones whose joint contribution in label prediction is significant. To formalize this idea, we introduce the notion of the power of classification of a subset, say $K$, of features, defined below:
\begin{defi}[Power of classification of feature subset $K$, $P_{SV}(K)$] 
Given a training dataset $D$ of size $m$ with feature values $\{x_{i1},\ldots,x_{in}\}_{i=1}^m$ and labels $\{y_i\}_{i=1}^m$, the power of classification of a set of features $K =\{j_1,j_2,\ldots,j_k \} \subset N$ is defined as follows:
\begin{equation}
P_{SV}(K) = \frac{\sum\limits_{i=1}^{m_{te}}\mathbf{1}_{[y_if_K^*(x_{ij_1},x_{ij_2},\ldots,x_{ij_k}) \geq 0]}}{\sum\limits_{i=1}^{m_{te}}\mathbf{1}_{[y_if_N^*(x_{i1},x_{i2},\ldots,x_{in}) \geq 0]}},
\end{equation}
where $f_K^*(\cdot)$ and $f_N^*(\cdot)$ are the optimal  {linear} classifiers in the respective subspaces and 
$m_{te}$ (different from $D$) is the number of sample points used for testing the classifiers and  $\mathbf{1}_{[A]}$ is the indicator function with value 1 if $A$ holds; else has 0 value. 
\end{defi}

The higher the value of $P_{SV}(K)$, the higher is the joint influence of the subset $K$ in classification. 
{The powerful subset $K = SVEA_{neg}$ is not pre-decided but determined by SVEA. Due to Shapley value's property of identifying the important players based on their contributions, SVEA scheme identifies features that play a dominating role in the task of classification and forms a set $K$.}
We demonstrate this  FSS phenomenon using Synthetic dataset 2 ($SD2$) in SM D.2. 
We give details about the  FSS interpretation for UCI datasets with SVEA, $\{e_j(m)\}_{j\in N}$ being negative in Section \ref{subsec: real_neg_SV_exp_feature}. 
Besides, we observe as in SM D.5  that $l_1$-regularized squared hinge loss based ERM doesn't identify important features for UCI datasets like Heart, Pima diabetes, and Thyroid.
\subsection{Relation between negative valued  SVEA  and dimension reduction} \label{subsec: neg_dr}
By dimension reduction from $\mathbb{R}^n$ to $\mathbb{R}^d$, we mean that $(n-d)$ feature dimensions have zero class conditional expected values and minimal class conditional variance. 
We consider a special structure on the distribution for dimension reduction and provide the following result:
\begin{thm}\label{thm: dimen_gen}
Consider the random variables $\mathbf{X}\subseteq \mathbb{R}^n,~Y\in \{-1,1\}$. Let $A :=\{ k \in N : E[X_k|Y=y]=0, var(X_k|Y=y) \leq \epsilon^{\prime}_k, ~\mathbb{P}[X_k \cap X_{k^{\prime}}|Y=y] = \mathbb{P}[X_k|Y=y]\times \mathbb{P}[X_{k^{\prime}}|Y=y],~ \forall k\neq k^{\prime} \in N,~ y\in Y\}.$ Also, $X_j, ~j\in A^c$ are independent of $X_k, k\in A.$ Then, 
\begin{equation} \label{eq: dimen_red_formalize}
\mathbb{P}[(X_j,X_k): X_j \in \mathbb{R}, |X_k| \leq \epsilon_k, j \in A^c, k \in A] \geq \prod_{k\in A} \left(1- \frac{var(X_k|Y=y)}{\epsilon_k^2} \right).
\end{equation}
\end{thm}
Proof of Theorem \ref{thm: dimen_gen} is available in SM A.4. Consider the R.H.S of Eq. \eqref{eq: dimen_red_formalize}. Since, $Var(X_k|Y)$ is very small, the cross product terms in the expansion of R.H.S can be ignored. Hence, using the first order approximation provides a high probability ($1-\delta$) bound with $\delta \approx \sum\limits_{k \in A}\frac{Var(X_k|Y)}{\epsilon^2_k}.$ The above analysis says that the dataset lies in the lower sub-space whose basis corresponds to features with indices belonging to $A^c$ with high probability. We observed that in such datasets, $e_j(m)<0$ for feature $X_j, ~j \in A^c.$ Hence, the lower dimensional subspace has basis corresponding to features $X_j$ with SVEA $e_j(m) < 0$. We consider a $2$-dimensional synthetic dataset (generated using the technique given in \cite{efron1997improvements}) to demonstrate the above aspect.


\begin{figure}[h!]
\centering
\includegraphics[scale=0.45]{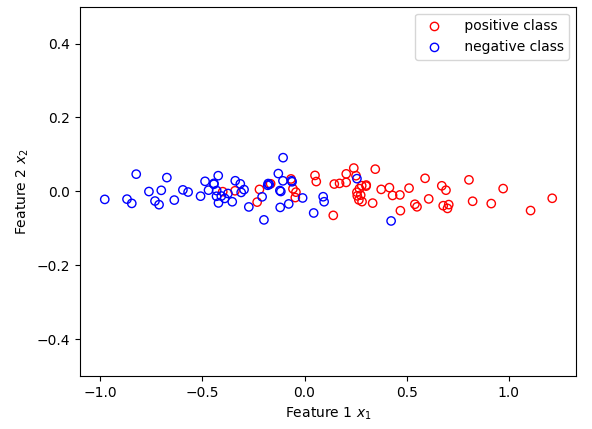}
\caption{\footnotesize{Scatter plot of 50 data points from  SD1. Here, $P(\mathbf{X}=(X_1,X_2):X_1\in \mathbb{R}, |X_2|\leq 0.03) \geq 0.99$,  i.e., the data predominantly lies in 1-d subspace corresponding to feature 1 with Shapley value based error apportioning $e_1(m)<0$.}}
\label{fig:dim_red_2d}
\end{figure}

\textbf{Synthetic dataset 1 (SD1):} We first generate $1000$ binary labels $Y$ uniformly at random and then,  a $2$-dimensional feature vector $X$ for each label by drawing a sample such that  $X|Y=1 \sim N([0.3; 0], [0.1, 0; 0, 0.001])$ and $X|Y=-1 \sim N([-0.3;0], [0.1, 0; 0, 0.001])$. SVEA $e_j(m), ~ j=\{1,2\}$ for 2 features is $[-0.073;  0.494]$. Invoking Theorem \ref{thm: dimen_gen}, the probability of the event that feature 2's value is within an $\epsilon$-ball, i.e., $P(|X_2| \leq \epsilon)$ is greater than $\left(1- \frac{var(X_2|Y=y)}{\epsilon^2}\right)$. 
When $\epsilon = 0.03$, we have $|X_2|\leq 0.03$ with probability more than $0.99$ as evident in Figure \ref{fig:dim_red_2d}. Also, using SVM, 1-dimensional classifier (trained on feature 1 only) has test accuracy of $0.85$ which is same as that of 2-dimensional classifier (test accuracy $= 0.855$). Hence, SD1  is reducible to a subspace with basis as the feature ($1$ here) with $e_1(m)<0$. In SM D.2, we provide one more Synthetic dataset 3 ($SD3$) example where the 6-dimensional dataset is predominantly lying in a 3-dimensional subspace as identified by SVEA.

We would like to emphasis that this is different from Principal Component Analysis (PCA) in two aspects: firstly, unlike our method, PCA does not use the labels, and secondly, the one-to-one mapping between the transformed dimensions (reduced) and the features are well defined in our scheme but not in PCA. 
\subsection{Identification of non-important features for classification by SVEA scheme} \label{subsec: eta_error_decomp}
In this section, we attempt to understand what contributes to the excess $0$-$1$ risk of surrogate loss function based classifiers. In this direction, we first consider a simple distribution $\mathcal{D}$ and present an explicit form of in-class probability $\eta(\mathbf{x})$. Following lemma is a special case of result provided in \cite{tripathi2019cost}. For the sake of completeness, a proof of Lemma  \ref{lem: eta-normal} is available in SM A.5.
\begin{lem} \label{lem: eta-normal}
Let Y has Bernoulli distribution with parameter $p$. Let $\mathbf{X} \subset \mathbb{R}^n$  be such that $\mathbf{X}|Y = 1 \sim N(\boldsymbol{\mu}_+,\Sigma)$ and $\mathbf{X}|Y = -1 \sim N(\boldsymbol{\mu}_-,\Sigma)$. Let $A_e =\{$ even numbers between 1 to n $\}$ and $A_o = \{\text{odd numbers between 1 to n}\}$. Also, $\mu_{-,j} = \mu_{+,j}~~ \text{if } j\in A_e$, $\mu_{-,j} = -\mu_{+,j}~~ \text{if } j\in A_o$ and $\Sigma = a\mathbf{I}_{n\times n}, ~ a>0.$ Then, the in-class probability $\eta(\mathbf{x}) = P(Y=1|\mathbf{X}=\mathbf{x})$ is given as follows:
\begin{equation}\label{eq: eta-normal}
\eta(\mathbf{x}) = \left[1 + \frac{1-p}{p}\exp\left(-\frac{2}{a}\left( \sum\limits_{j\in A_0}x_j\mu_{+,j}\right)\right)\right]^{-1}.
\end{equation}
\end{lem}
This implies that the Bayes classifier $f^*_{\text{0-1}}(\mathbf{x}) = sign(\eta(\mathbf{x})-1/2) \in \mathcal{H}$ doesn't depend on feature $X_j, ~j \in A_e.$
We observe that SVEA scheme is able to identify those features which do not appear in the Bayes classifier as the ones with $e_j(m)>0.$  Computational evidence for this claim is available in SM D.2 using Synthetic dataset 4 (SD4). This observation has following implication: finite sample and surrogate loss based classifiers contribute to the excess $0$-$1$ risk because coefficients of features of which $\eta(\mathbf{x})$ is not a function, are non-zero and for such features $e_j(m)>0.$ 

To support our observation, we first provide a detailed interpretation of the decomposition of excess $0$-$1$ risk of a surrogate loss function $l$ based classifier $\hat{f_l} = \arg\min\limits_{f\in \mathcal{H}_{lin}} \hat{R}_{l}(f):=\frac{1}{m}\sum\limits_{i=1}^{m}l(\mathbf{x_i},f(\mathbf{x_i})).$ Let $f^*_{l}\in \mathcal{H}$ denote the minimizer of true $l$-risk, $R_{l}(f):= E_{\mathcal{D}}[l(\mathbf{x},f(\mathbf{x}))]$. Also, empirical $0$-$1$ risk is  $\hat{R}_{\text{0-1}}(f):=\frac{1}{m}\sum\limits_{i=1}^{m}l_{\text{0-1}}(\mathbf{x_i},f(\mathbf{x_i})).$
Consider the excess $0$-$1$ risk of $\hat{f}_l$ as given below:
\begin{scriptsize}
	\begin{eqnarray} \nonumber
	R_{\text{0-1}}(\hat{f}_l) - R_{\text{0-1}}(f^*_{\text{0-1}}) &=& R_{\text{0-1}}(\hat{f}_l) -R_{\text{0-1}}(f^*_{l}) + R_{\text{0-1}}(f^*_{l}) - R_{\text{0-1}}(f^*_{\text{0-1}}) \\ \nonumber
	&=& {(\hat{R}_{\text{0-1}}(\hat{f}_l) - \hat{R}_{\text{0-1}}(f^*_{l}))} + {(R_{\text{\text{0-1}}}(\hat{f}_l) -\hat{R}_{\text{0-1}}(\hat{f}_{l}) + \hat{R}_{\text{0-1}}(f^*_l) -R_{\text{0-1}}(f^*_{l}))} \\ \nonumber
	& & + {(R_{\text{0-1}}(f^*_l) -R_{\text{0-1}}(f^*_{\text{0-1}}))}\\ \nonumber
	&=& \underbrace{(\hat{R}_{\text{0-1}}(\hat{f}_l) - \hat{R}_{\text{0-1}}(f^*_{\text{0-1}}))}_{\text{Error }1} + \underbrace{(R_{\text{0-1}}(\hat{f}_l) -\hat{R}_{\text{0-1}}(\hat{f}_{l}) + \hat{R}_{\text{0-1}}(f^*_l) -R_{\text{0-1}}(f^*_{l}))}_{\text{Error }2} \\ \label{eq: decom_error}
	& & + \underbrace{(R_{\text{0-1}}(f^*_l) -R_{\text{0-1}}(f^*_{\text{0-1}}))}_{\text{Error }3}.
	\end{eqnarray}
\end{scriptsize}

The last equality follows from the fact that for classification calibrated loss functions $sign(f^*_l)=sign(f^*_{\text{0-1}})$ \cite{zhang2004statistical,Bartlett2006,steinwart2008support} and hence their $0$-$1$ risks (both empirical and true) are equal. This also implies that $\text{Error }3$ is zero. $\text{Error }2$ is the error due to the use of a sample based risk instead of true risk. This term will vanish as the sample size increases. Finally, we have $\text{Error }1$ which is a combination of using a finite sample, using a surrogate loss function and restricting the hypothesis class to say linear class $\mathcal{H}_{lin}.$ After identifying the important features based on the sign of SVEA $e_j(m)$, one can get some extra insight about $\text{Error } 1$'s decomposition. We believe that $\text{Error }1$ will also have error due to the following reason: in spite of $f^*_{\text{0-1}}(\mathbf{x})$ not depending on features $X_j,~j\in A_e$, $\hat{f}_l$ can be a function of features $X_j,~j\in A_e$ (apart from those features $X_j,~j\in A_o$). 
This error could either be an additional component in $\text{Error }1$ or absorbed  in the existing components explained before.

For surrogate loss functions, we consider hinge loss (using SVM), logistic loss (using Logistic regression) and exponential loss (using ExpERM given in \cite{ExpERM18}). It was observed that none of the surrogate loss functions considered had zero coefficient corresponding to features $X_j,~j\in A_e$ in a linear classifier. This implies that inclusion of such features is one of the contributors to the extra error (excess risk) incurred by the surrogate loss function based classifiers in comparison to Bayes classifiers, in particular, to $\text{Error }1$ term. Details about these results are provided in SM D.2 using Synthetic dataset 4 $(SD4).$


\subsection{SVEA based decomposition of true hinge risk among features} \label{subsec: converg_SV_error}
In this section, we attempt to understand what one can say about the nature of SVEA, $\{e_j(m)\}_{j\in N}$ as sample size $m$ increases. In various computational experiments, we observed that the Shapley values of the classification game, $\{\phi_j(m)\}_{j\in N}$ and SVEA, $\{e_{j}(m)\}_{j\in N}$, converge to limiting values, say $\{\phi_j\}_{j\in N}$ and $\{e_{j}\}_{j\in N}$ as the sample size $m$ increases. Since, $\{e_j(m)\}_{j\in N}$ is the empirical hinge risk of feature $j$, we can interpret the limiting value $\{e_j\}_{j\in N}$ as the true hinge risk of feature $j\in N$. In other words, $\{e_j\}_{j\in N}$ represents the components in the decomposition of minimum true hinge risk $R_{l_{hinge}}(f^*_{hinge})$, i.e., $\sum_{j\in N}e_j = R_{l_{hinge}}(f^*_{hinge})$ where $R_{l}(f):= E_{\mathcal{D}}[l(\mathbf{x},f(\mathbf{x}))]$.
As $\{e_j(m)\}_{j\in N}$ is computed for a given sample, it can be interpreted as an estimate of the true hinge risk of feature $j\in N$. Also, since it converges to $\{e_j\}_{j\in N}$ as sample size $m$ increases, one can expect it to be a consistent estimate of $\{e_j\}_{j\in N}$. To illustrate this convergence, we provide an example based on a 7-dimensional Synthetic dataset 5 $(SD5)$ in SM D.2.

\subsection{Stable SVEA and convex classification games}\label{subsec: stable_app}
In this section, we discuss additional properties of the allocation measure of classification games, in terms of stability and convexity of the games.
Core, a well-known solution concept in cooperative game theory, is defined as the set of allocations in the imputation set (allocations satisfying IR and collective rationality; details in SM B.3), which are Coalitional Rational (CR). Such allocations are said to be stable as no player or group of players objects to the allocations in the core. As we are interested in apportioning of total training error, CR is desirable; so, we extend the notion of the core to SVEA, denoted by $C_{E}(m)$, as follows:
\begin{equation*}
\begin{scriptsize}
    C_{E}(m) :=  \{\mathbf{e}^{\prime}(m) \in \mathbb{R}^n \vert \sum_{j \in S}e^{\prime}_j(m) \leq tr\_er(S,m), ~ \forall S \subseteq N, \sum_{j \in N}e^{\prime}_j(m) = tr\_er(N,m)\}.
    \end{scriptsize}
\end{equation*}
Allocations belonging to $ C_E(m)$ will also be stable in the sense that any feature or group of features will not object to the error allocations $e^{\prime}_j(m),~ j \in N$. So, we would like our SVEA to be an element of  $C_E(m)$. However, as seen in Table 2 of SM B, whether $SVEA \in C_E(m)$ or not depends on the dataset;  IR and Collective rationality holds, but, CR, {\footnotesize $\sum_{j \in S} e_j(m) \le tr\_{er}(S,m)$} may not hold for a given dataset. 

If a game is convex (definition in SM B), then its Shapley value belongs to the core. As SVEA is an affine transformation of Shapley value, one would expect that for such games, SVEA belongs to core's counterpart $C_E(m)$; true for Haberman dataset as seen in Table 2 of SM B. Also, for the datasets (Thyroid, Pima, Magic, and Banknote) in which $e_j(m)$ belongs to $C_E(m)$, there are some features for which these $e_j$s are negative;  
note that such features are unique for a given sample of a dataset (as Shapley value is unique for a sample).  
As observed in Section \ref{subsec: neg_fss}, classifiers based on these features alone can yield the accuracy comparable to those obtained with full feature set ($P_{SV} \approx 1$)  and such a unique set of features with allocations also in $C_E(m)$ have stable allocations.

\subsection{Sample bias robustness of SVEA scheme} \label{subsec: sample_bias_robust_tech}
With the goal of being robust to sample bias, we provide interval estimates for SVEA of features by using multiple sub-samples from a given dataset. A feature's joint contribution in label prediction is significant if the interval estimate of SVEA for a feature lies on the left of origin on $\mathbb{R}$. The procedure is first to partition the dataset into multiple disjoint sub-samples and compute the apportioning for each sub-sample. Then, a group of 30 such sub-samples is selected, and using CLT, the average apportioning $\bar{e}_j^g, j \in N$ for each group $g$ is asymptotically normally distributed with unknown mean $\mu_e$ and variance $\sigma_e^2$. Next, using $\bar{e}_j^g$, we compute t-distribution based $100(1-\alpha)$ confidence intervals. By the definition of confidence intervals, we have following high probability statement:
\begin{equation*}
    P(e_j^p \in [\bar{\bar{e}}_j \pm t^*_{\alpha/2, G-1} (s_j/ \sqrt{G})]) \geq 1-\alpha, ~~\forall j \in N,
\end{equation*}
where $e_j^p$ is the population mean for the error apportioning of feature $j \in N$, $\bar{\bar{e}}_j = \frac{1}{G}\sum_g \bar{e}_j^g$ and $s_j = (\frac{1}{G-1}\sum_g (\bar{e}_j^g - \bar{\bar{e}}_j)^2)^{1/2}$ and $t^*_{\alpha/2, G-1}$ is the upper $\alpha/2$ critical value for the $t$ distribution with $G-1$ degrees of freedom ($G$ is the number of groups). More details of this procedure are provided in SM D.1.
Based on our experiments in Section \ref{subsec: real_SVNeg_exp_inter}, we observe that the interval estimates also lead to the same threshold of $0$ while performing FSS.
Also, the conclusions are robust to sample bias due to multiple averaging. This shows that the behaviour of features with SVEA<0 mentioned in Section \ref{subsec: stable_app} is a property of the dataset and not of a particular sample. In  SM D.2, we implement the above technique on Synthetic dataset 4 (SD4) and observe that the results are consistent with Section \ref{subsec: eta_error_decomp}.
{The above presented method is tailor-made for SVEA scheme. A more general framework to address the instability issue, i.e., change in sample leading to change in feature subset is presented in \cite{nogueira2017FSSstability}. They first  show that any  existing stability measure doesn't possess all five desirable properties which a stability measure should have. Then, taking a statistical approach they propose a novel measure which is treated as an estimator of true stability. Note that the ``stability of a feature selection algorithm'' considered in this sub-section is different from the ``stability of an allocation'' in Section \ref{subsec: stable_app}.}


\section{Computational experiments} \label{sec: Computations} 
In this section, we empirically demonstrate the implications of SVEA being negative for some features on real-world datasets from \cite{alcala2011keel,UCI_dataset}. For the FSS interpretation, we train a classifier using SVM (linear and rbf kernel), Logistic Regression (LR), Random Forest (RF) and Multi Layer Perceptron (MLP) to compute $P_{SV}$. 
We also compare our SVEA approach to  Recursive Feature Elimination with Cross-Validation (RFECV) and RefliefF. Implementation of RFECV with $5$-folds was done in Scikit learn module of Python \cite{scikit-learn}. For ReliefF, we used the implementation of \cite{li2016feature_reliefF} with neighbour parameter $k=2$. 
All the above mentioned algorithms are implemented in { \em Python 3} with { \em Gurobi 8.0.0} solver for LPs, on a machine equipped with 4 Intel Xeon 2.13 GHz cores and 64 GB RAM. 
To account for randomness, we repeat each experiment 5 times and report the average test accuracy (and standard deviation).

\textbf{Real datasets: Demonstration of FSS using $\mathbf{P_{SV}}$ with threshold 0 on SVEA values} \label{subsec: real_neg_SV_exp_feature}
Out of the 14  benchmark dataset considered, 3 datasets namely, WDBC, Banknote and Iris0 are almost linearly separable. We used Breastcancer, German and Thyroid datasets from (\url{http://theoval.cmp.uea.ac.uk/matlab}).
First, we consider those UCI datasets  in which some features have negative SVEA and compute their Power of classification, $P_{SV}.$
Let $SVEA_{neg}$ be the set of features  with $e_j(m)<0$ and $P_{SV}(SVEA_{neg})$ is the power of classification of the set $SVEA_{neg}$. From Table \ref{Table:real_dataset_SV_neg}, the value of $P_{SV}(SVEA_{neg})$ is close to 1 in most of the cases. Hence, the features in the set $SVEA_{neg}$ have a large joint contribution towards classification. Five different types of classifiers support this behavior, viz., SVM (linear), LR, RF, SVM(rbf) and MLP. Except for the Thyroid dataset, the $P_{SV}$ value for a given dataset is similar across classifiers. Moreover, we have also computed the $P_{SV}$ for all subsets of the $SVEA_{neg}$ set for Heart dataset and observed that in comparison to its subsets, $SVEA_{neg}$ has the highest value of $P_{SV}$. For heart dataset, we observed that average accuracy ($\pm$ s.d.) over three trials with only feature 3 is $0.765\pm 0.046$, with only feature 12 is $0.697\pm0.043$, with only feature 13 is $0.77\pm 0.453$, with features 3 and 12 is $0.771\pm 0.046$, with features 12 and 13 is $0.77 \pm 0.045$ and with features 3, 12 and 13, i.e., $SVEA_{neg}$ set is $0.81 \pm 0.008$ which is highest among all the subsets. 

In addition to the datasets reported in 
Table \ref{Table:real_dataset_SV_neg}, we also implemented our SVEA scheme on other large scale datasets like EEG-Eye state dataset (14 features, 14980 examples), Numerai dataset (21 features, 96320 examples). However, we did not observe any feature with negative value of SVEA and hence in Table \ref{Table:real_dataset_SV_neg} we only report datasets that had features with SVEA values less than zero.

\begin{figure*}[!h]
    \centering
    \begin{subfigure}[b]{0.475\textwidth}
        \centering
        \includegraphics[width=1.04\textwidth]{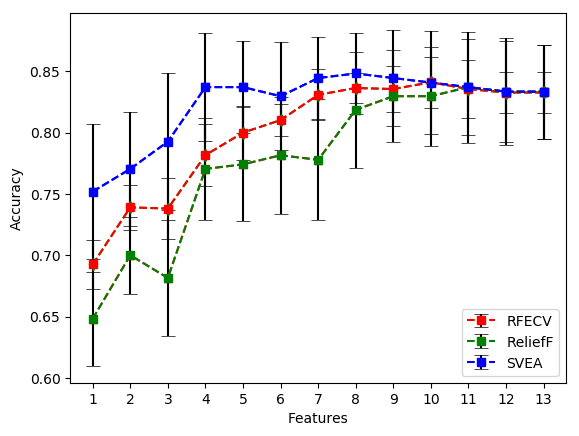}
        \caption{\footnotesize{Dataset: Heart(270,13)}}
        {}    
    \end{subfigure}
    \hfill
    \begin{subfigure}[b]{0.475\textwidth}  
        \centering 
        \includegraphics[width=1.04\textwidth]{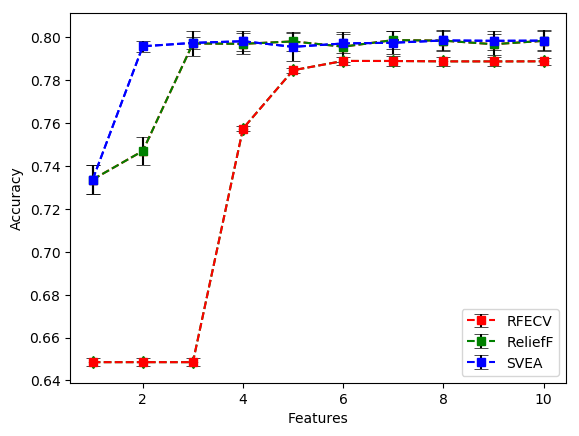}
        \caption{\footnotesize{Dataset: Magic(19020,10)}}
        {}       
    \end{subfigure}
    \vskip\baselineskip
        \begin{subfigure}[b]{0.475\textwidth}
        \centering
        \includegraphics[width=1.04\textwidth]{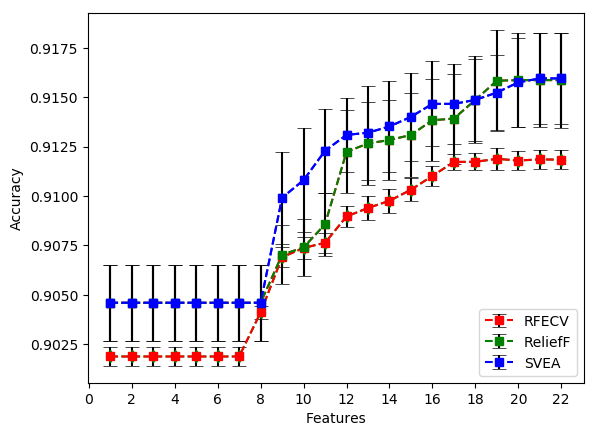}
        \caption{\footnotesize{Dataset: IJCNN(35000(tr),22)}}
        {}    
    \end{subfigure}
    \hfill
    \begin{subfigure}[b]{0.475\textwidth}   
        \centering 
        \includegraphics[width=1.04\textwidth]{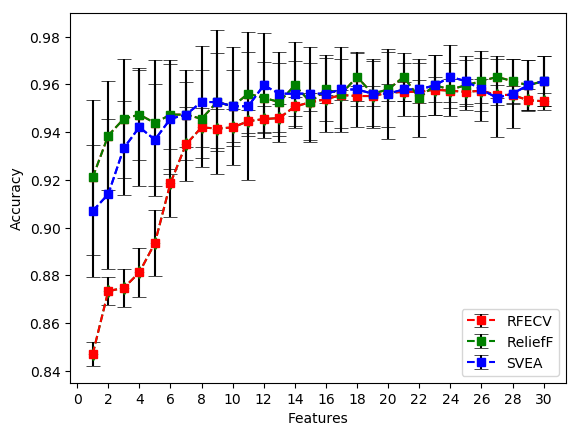}
        \caption{\footnotesize{Dataset: Wdbc(569,30)}}
        {}       
    \end{subfigure}
    \caption{\footnotesize Plot of test accuracy vs number of features used to train the linear classifier using SVM. For each scheme, we have $95\%$ error bar computed over 5 iterations. Given a fixed number of features and a lower bound on accuracy, SVEA provides the feature subset which leads to highest test accuracy in most of the cases. Number of trials in the plot for Magic dataset is three; for other datasets number of trials is five.}
    \label{fig: comparison_real_data}
\end{figure*}

\textbf{Real datasets: Demonstration of FSS with a user given threshold and comparison to RFECV and ReliefF} {For a user given  feature size, say $l$, with due justification in terms of SVEA, our scheme can identify the $l$ sized feature set with best test accuracy based on the  ranking of the SVEA values.} We demonstrate this property of SVEA scheme and compare it to RFECV and ReliefF. 
 To do this, we order the features based on the score/SVEA value for each scheme and  then plot the SVM test accuracy of linear classifiers learnt using first $l$ features (Figure \ref{fig: comparison_real_data}).
Clearly, too few features leads to degradation in performance, and too many features defeat the purpose of feature selection. In comparison to other methods, our scheme achieves the highest accuracy when one looks for a trade-off by selecting a subset of features whose cardinality is neither too small nor too large. {Also, if the user given threshold on the number of features is $l$, then SVEA has best accuracy as observed in Figure \ref{fig: comparison_real_data} for Magic, Heart and IJCNN dataset with $l=2,3,5$ respectively.}

{Using a statistical significance test to compare our scheme to RFECV and ReliefF is not straight forward due to computation of incremental feature accuracy, so we use the measure that given a fixed number of features and a lower bound on accuracy, a good scheme should identify feature subset leading to high accuracy. However, for the sake of completeness, we still performed many Friedman tests (using Scikit-posthocs package in python)  by fixing the number of features across datasets and found no significant difference between the schemes at 5\% level of significance except for the cases where SVEA is better as found by Nemenyi posthoc test \cite{demvsar2006statisticalTests}.}
Comparative plots for other datasets and additional explanations (Table 15) are provided in SM D.5. Using Pima Diabetes dataset, we show that the high value of $P_{SV}(SVEA_{neg})$ in Table \ref{Table:real_dataset_SV_neg} is a case of FSS and not of dimension reduction. Details of latter argument are presented in Table 13 of SM D.5.

\begin{figure*}[!h]
    \centering
    \begin{subfigure}[b]{0.475\textwidth}
        \centering
        \includegraphics[width=1.04\textwidth]{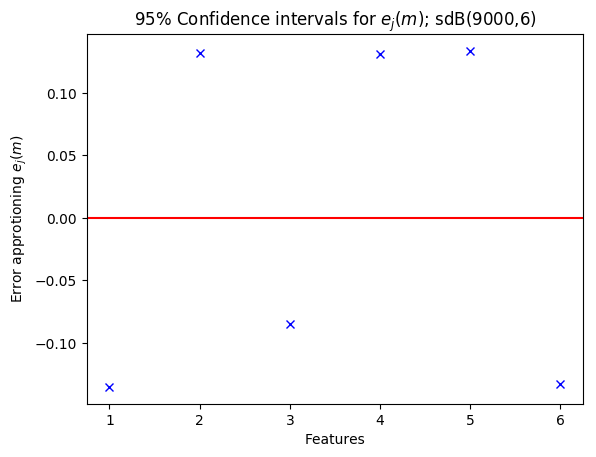}
        \caption*{}%
        {}    
    \end{subfigure}
    \hfill
    \begin{subfigure}[b]{0.475\textwidth}  
        \centering 
        \includegraphics[width=1.04\textwidth]{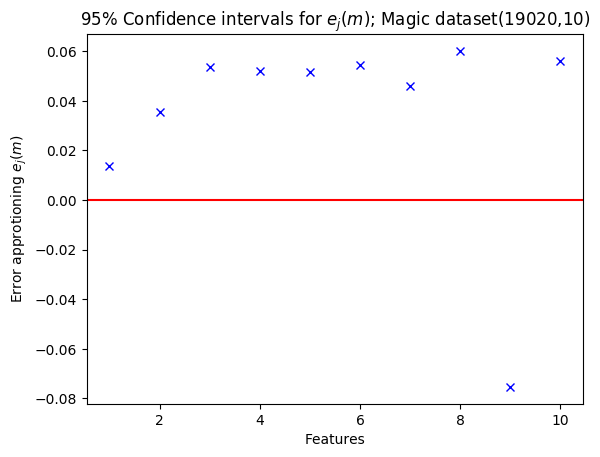}
        \caption*{}%
        {}       
    \end{subfigure}
    \vskip\baselineskip
        \begin{subfigure}[b]{0.475\textwidth}
        \centering
        \includegraphics[width=1.04\textwidth]{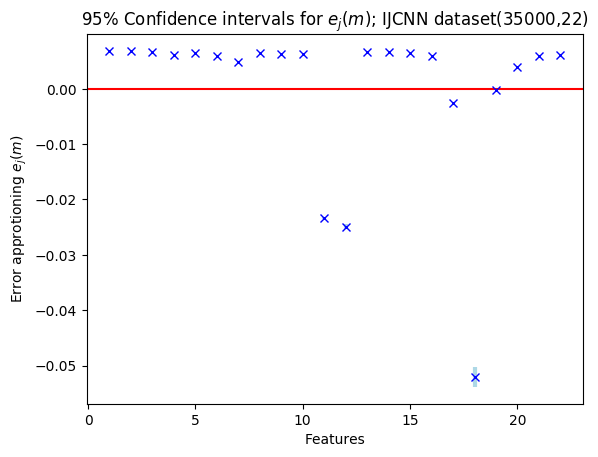}
        \caption*{}%
        {}    
    \end{subfigure}
    \hfill
    \begin{subfigure}[b]{0.475\textwidth}   
        \centering 
        \includegraphics[width=1.04\textwidth]{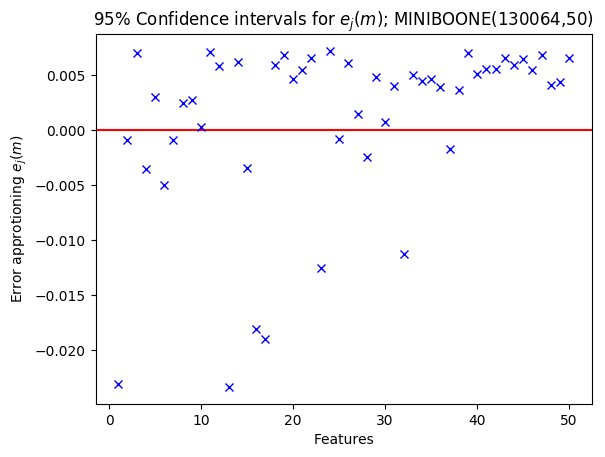}
        \caption*{}%
        {}       
    \end{subfigure}
    \caption{\footnotesize Above plots depict $95\%$ confidence intervals for SVEA of features for 6 dimensional synthetic dataset sdB and 3 UCI datasets. As the importance of a feature is based on intervals of SVEA, we can say that on an average the population value of SVEA would lie inside the interval estimates 95 out of 100 times (which are below $0$) and hence important features via SVEA is a pattern manifestation of underlying population.} 
    \label{fig: Syn_Real_datasets_interval_estimates}
\end{figure*}

\textbf{Real datasets: Sample bias robust interval estimates} \label{subsec: real_SVNeg_exp_inter}
In this section, we demonstrate our idea of addressing the issue of sample bias while making conclusions based on the SVEA scheme. Since the technique requires partitioning the whole dataset into many disjoint subsets, large sample sized datasets are considered. Figure \ref{fig: Syn_Real_datasets_interval_estimates} shows t-distribution based $95\%$ confidence intervals of SVEA estimates for synthetic dataset SD4 (sdB) and real datasets Magic, IJCNN (\cite{LIBSVMdatsets}) and MINIBOONE (\cite{OpenML2013}). There is a partitioning of feature set into two subsets; one in which the features have their SVEA's confidence intervals above origin and other in which the features have their SVEA's confidence intervals below the origin. As $P_{SV}(SVEA_{neg})$ (given in Table \ref{Table:real_dataset_SV_neg}) for the latter subset of features is high, one can conclude that these features have a large contribution in label prediction. Since the feature set partitioning is based on interval estimates, the conclusions regarding important features are robust to sample bias.

\begin{table}[h!]
\centering
\setlength{\tabcolsep}{1.5pt}
\renewcommand{\arraystretch}{1.2}
{\small
	\begin{tabular}{|c|c|c|c|c|c|}
		\hline
		\textbf{\begin{tabular}[c]{@{}c@{}}Dataset\\(m,n)\end{tabular}} &  \textbf{Clf} & \textbf{\begin{tabular}[c]{@{}c@{}}Avg Acc \\ ($\pm$std dev)\end{tabular}} & \textbf{$SVEA_{neg}$} & \textbf{\begin{tabular}[c]{@{}c@{}}Avg Acc ($\pm$std dev) \\  $SVEA_{neg}$\end{tabular}} & \textbf{\begin{tabular}[c]{@{}c@{}}$P_{SV}$\\ \end{tabular}} \\ \hline
		\multirow{3}{*}{\textbf{\begin{tabular}[c]{@{}c@{}}Thyroid \\ (215,5)\end{tabular}}} & SVM & 0.89$\pm$0.0145 & \multirow{3}{*}{$\{4\}$} & 0.83$\pm$0.0087 & 0.93\\ \cline{2-3} \cline{5-6} 
		&  LR & 0.89$\pm$0.0315 &  & 0.79$\pm$0.0328  & 0.89  \\ \cline{2-3} \cline{5-6} 
		&  RF & 0.93$\pm$0.0400 &  & 0.76$\pm$0.0641 & 0.82 \\ \cline{2-3} \cline{5-6} 
		& SVM\_K & 0.95+-0.025 &  & 0.78+-0.0405 & 0.82 \\ \cline{2-3} \cline{5-6} 
		& MLP & 0.94 &  & 0.78+-0.0372 & 0.83  \\ \hline
		\multirow{3}{*}{\textbf{\begin{tabular}[c]{@{}c@{}}Pima\\ Diabetes\\ (768,8)\end{tabular} }} & SVM & 0.77$\pm$0.0060 & \multirow{3}{*}{$\{2\}$} & 0.75$\pm$0.008 & 0.98 \\ \cline{2-3} \cline{5-6} 
		&  LR & 0.75$\pm$0.0075 &  & 0.73$\pm$0.0218 & 0.97 \\ \cline{2-3} \cline{5-6} 
		&  RF & 0.73$\pm$0.0097 &  & 0.72$\pm$0.0230 & 0.98 \\ \cline{2-3} \cline{5-6} 
		& SVM\_K & 0.72+-0.0257 &  & 0.71+-0.0544 & 0.99 \\ \cline{2-3} \cline{5-6} 
		& MLP & 0.67+-0.0269 &  & 0.66+-0.0273 & 0.98  \\ \hline
		\multirow{3}{*}{\textbf{\begin{tabular}[c]{@{}c@{}}Magic \\ (19020,10)\end{tabular}}} & SVM & 0.79$\pm$0.0058  & \multirow{3}{*}{$\{9\}$} & 0.74$\pm$0.0060 & 0.93 \\ \cline{2-3} \cline{5-6} 
		&  LR & 0.79$\pm$0.0058 &  & 0.73$\pm$0.0061 &  0.92\\ \cline{2-3} \cline{5-6} 
		&  RF & 0.75 $\pm$ 0.0035 &  & 0.73$\pm$ 0.0097 & 0.97 \\ \cline{2-3} \cline{5-6} 
		& SVM\_K & 0.82+-0.0620 &  & 0.84+-0.0544 & 1.02 \\ \cline{2-3} \cline{5-6} 
		& MLP & 0.82+-0.007 &  & 0.73+-0.0075 & 0.89 \\ \hline
		\multirow{3}{*}{\textbf{\begin{tabular}[c]{@{}c@{}}Heart \\ (270,13)\end{tabular}}}  & SVM & 0.84$\pm$0.0478 & \multirow{3}{*}{$\{3,12,13\}$} & 0.81$\pm$0.0080 & 0.96 \\ \cline{2-3} \cline{5-6} 
		&  LR & 0.82$\pm$0.0381 &  & 0.79$\pm$0.0309 &  0.96\\ \cline{2-3} \cline{5-6} 
		&  RF & 0.82$\pm$0.0343 &  & 0.84$\pm$0.0578 & 1.02 \\ \cline{2-3} \cline{5-6} 
		& SVM\_K & 0.64+-0.0608 &  & 0.84+-0.0544 & 1.31 \\ \cline{2-3} \cline{5-6} 
		& MLP & 0.77+-0.0482 &  & 0.79+-0.0035 & 1.02 \\ \hline
		\multirow{3}{*}{\textbf{\begin{tabular}[c]{@{}c@{}}IJCNN\\(126701,22)\end{tabular}}} & SVM & 0.91 & \multirow{3}{*}{\begin{tabular}[c]{@{}c@{}}$\{11,12,17,$\\ $18,19\}$\end{tabular}} & 0.90 & 0.99 \\ \cline{2-3} \cline{5-6} 
		&  LR & 0.91 &  & 0.90 &  0.99\\ \cline{2-3} \cline{5-6} 
		&  RF & 0.90 &  & 0.90 & 1 \\ \cline{2-3} \cline{5-6} 
		&  SVM\_K& 0.979 &  & 0.957 & 0.98  \\ \cline{2-3} \cline{5-6} 
		&  MLP & 0.977 &  & 0.956 & 0.98 \\ \hline
\end{tabular}}
\caption{Accuracies of the datasets having negative SVEA for features in $SVEA_{neg}$ for SVM (linear kernel), LR, RF, SVM\_K (rbf kernel) and single layer MLP classifier. The sixth column has the accuracy of the classifiers learnt  only on features in $SVEA_{neg}$. $P_{SV}$ is the ratio of accuracies in column 3 and column 5 when the important feature subset is $(SVEA_{neg})$. SVM, LR, SVM\_K parameter $C \in \{0.1,1,50,500\}$, RF parameters $n\_estimators \in \{0.1,1,50,500\}$ and $max\_depth = 2$, SVM\_K parameter $\gamma = \frac{1}{n*Var(X)}$, MLP regularization parameter $\alpha \in \{10^{-7},\ldots,10^{-1}\}$, constant learning rate of $10^{-3}$, Relu activation and Adam solver. $m$ is the sample size and $n$ is the number of features.  No averaging is done for IJCNN as the train-test (35000+91701) partitioning is already available from the source. }\label{Table:real_dataset_SV_neg}
\end{table}

    
    

\section{Discussion} \label{sec: discussion} 
To interpret the influence of a single feature or their interactions on the label/class, we use the framework of transferable utility cooperative games and introduce a classification game with features as players and hinge loss based characteristic function (computed as LPs).
\subsection{Summary}
We propose SVEA scheme to apportion the total hinge loss based empirical risk among the features. As Shapley value is computationally expensive, we build on an approximation algorithm that does not compute the characteristic function (an LP) for all subsets at one go but only when needed.
From the perspective of classification, SVEA leads to the following contributions. 

\textit{Feature subset selection:} Features with negative SVEA are the ones whose joint contribution for label prediction is significant. \textit{Identification of lower-dimensional subspace:} Dataset lies in subspace with basis as features having negative SVEA value.
{Our scheme uses a universal threshold of $0$ on the SVEA value for all datasets to identify both the sets mentioned above. Also, our scheme can also identify the subset with the best accuracy based on ranking SVEA values if the feature set size is user-given.}
\textit{Decomposition of excess $0$-$1$ risk:} under a special structure on data distribution, only features (essential) with negative valued SVEA contribute to Bayes risk. However, we empirically observe that any surrogate loss based classifier depends on features non-important for classification also, and thereby increasing the excess $0$-$1$ risk of the classifier. We also provide an estimate of the unknown true hinge risk of each feature. 
We attempt to make our SVEA estimate robust to sample bias by computing interval estimates by averaging over multiple disjoint sub-samples.
We demonstrate all the above contributions and our  SVEA scheme compares favourably with the existing feature selection schemes RFECV and ReliefF on various synthetic and UCI datasets.

\subsection{Characteristic function $v(S,m)$ with regularization and  kernels} \label{subsec: kernel_reg_ntworking_exp}
Regularization to avoid over-fitting is a natural thing to do in most of the learning problems. We considered a $l_2$-regularized version of $tr\_er(S,m)$ defined in Section \ref{training_error_fn} and used in Eq.  \eqref{value_to_cost} for defining the characteristic function $v(S,m)$. However, this characteristic function turned out to be negative due to extra $\Vert \mathbf{w} \Vert^2$ term in $tr\_er(S,m)$. To avoid this issue, we defined the characteristic function $v_{reg}(S,m)$ (using regularized $tr\_er(S,m)$) which was empirically observed to be positive on all datasets. We compared the final results (for feature subset selection in Section \ref{subsec: neg_fss}) using $v(S,m)$ and $v_{reg}(S,m)$ and found that using regularization doesn't change important feature set $SVEA_{neg}$ (verified across 5 trials on 2 real and 3 synthetic datasets). Details with empirical results are available in SM D.3. 

We also considered the case when non-linear classifiers (via kernels) are used in the characteristic function. Note that regularization is necessary when using kernels to get the dot product of feature mapping $\phi(\mathbf{x})$. We used Eq. \eqref{value_to_cost} with kernelized and regularized $tr\_er(S,m)$ to obtain $v_{k,reg}(S,m)$. 
In the computational experiments, we observed variation in identification of important feature subset $SVEA_{neg}$ across trials, and in some cases, the important feature set (using $v_{k,reg}(S,m)$) did not have any common element with the set obtained using linear classifiers. Also, in most of the datasets, the test accuracy (with kernels) using all features did not improve over the linear case. One issue here is that $v_{k,reg}(S,m)$ is negative in most of the cases, and the trick used to make the characteristic function positive used in the linear case is not applicable here. The variation across trials could be attributed to the non-monotonic nature of $v_{k,reg}(S,m)$, which implies that the implicit assumption for Shapley value that the grand coalition will form is not satisfied. This issue is arising due to the use of regularization and exists in linear cases too. However, it is more prominent with kernels because here $v_{k,reg}(S,m)$ cannot be made positive. More details with supporting computational experiments are available in SM D.4. 

To summarize, in addition to the linear unregularized case that led to some interesting insights, we also considered two other cases, viz., linear regularized, and non-linear regularized. Even though the use of Shapley value in the linear regularized case could be justified, its performance is the same as that of linear unregularized and hence good. However, in the regularized kernel case, the performance of the SVEA scheme is not good as far as feature subset selection is considered; in fact, use of Shapley value can not be justified as the characteristic function is not monotone.

\subsection{Looking ahead}
A comparison of SVEA from $0$-$1$ loss function and other surrogate loss function based classification games would be interesting to explore; a ranking of surrogate losses can be expected. A thorough study on more game-theoretic aspects like understanding of dataset dependent properties of the game, modeling as NTU game, etc., could be another direction. Finally, in classification setup, a natural extension would be to generalize binary classification games to multi-class games.

In this paper, we have used linear and unregularized training error function, which makes sure that $v(\cdot,m)$ is monotonic (Proposition \ref{prop: game monotonic}) and use of Shapley value is justified. However, if some problem does require using regularization for feature subset selection, then our computational experience suggests that the characteristic function $v(\cdot,m)$ has to be suitably defined.

\nocite{Beleites2013samplesize}
\nocite{young1985monotonic}
\bibliographystyle{spmpsci}      
\bibliography{ref}   

	
\newpage
\appendix

\begin{center}
    \hrulefill 
    
   {\Large Supplementary material \\
    On feature interactions identified by Shapley values  \\ of binary classification games }
    
    \hrulefill
\end{center}

\section{Proofs and additional related work}
\subsection{Proof of Theorem \ref{thm: appor}} \label{app: appor}
\begin{proof}
From Eq. \eqref{eq: Shapley value}, the Shapely value for a player $j$ is given by 
\begin{equation*}
\phi_j(N,v(\cdot,m)) = \sum\limits_{S\subseteq N\backslash \{j\}} \frac{|S|!(n-|S|-1)!}{n!}[v(S\cup \{j\},m) - v(S,m)]. \\
\end{equation*}
Using efficiency axiom for Shapley value, we have
\begin{small}
\begin{eqnarray*}
    \sum_{j \in N} \phi_j(N,v(\cdot,m)) &=& v(N,m)\\
    \sum_{j \in N} \phi_j(N,v(\cdot,m))  &=& tr\_er(\{\emptyset\},m) - tr\_er(N,m) \\
\implies \hspace{1.5 cm}  tr\_er(N,m) &=& tr\_er(\{\emptyset\},m) -\sum_{j \in N} \phi_j(N,v(\cdot,m)) \\
    &=& \sum_{j \in N} \left( \frac{\tilde{c}(m)}{n} - \phi_j(N,v(\cdot,m)) \right),~~ \text{using LP in Section \ref{training_error_fn}. }
\end{eqnarray*}
\end{small}
Hence, the contribution of feature $j$ in the total training error is given as follows:
\begin{equation*}
    e_j(tr\_er(N,m)) = \frac{\tilde{c}(m)}{n} - \phi_j(N,v(\cdot,m)).
\end{equation*}
\qed
\end{proof}


\subsection{Proof of Theorem \ref{thm: apportion_train IR}}
\label{app: apportion_train IR}
\begin{proof}
In this theorem, we will show that the apportioning of the total training error satisfies individual rationality. The proof uses the definition of the apportioning of the total training error given in Eq. \eqref{eq: apport_total_train_error}. For all $j\in N$,
\vspace{-0.2 cm}
\begin{footnotesize}
\begin{eqnarray*}
    e_j(tr\_er(N,m)) &=& \frac{\tilde{c}(m)}{n} - \phi_j(N,v(\cdot,m)) \\
    &=& \frac{\tilde{c}(m)}{n} - \left[ \frac{1}{n} v(\{j\},m) + \sum\limits_{\substack{S\subseteq N\backslash \{j\}\\ S\neq \emptyset}} \frac{|S|!(n-|S|-1)!}{n!}[v(S\cup \{j\},m) - v(S,m)] \right] \\
    &=& \frac{\tilde{c}(m)}{n} - \left[ \frac{1}{n} (\tilde{c}(m)-tr\_er(\{j\},m)) \right] \\
    & & + \sum\limits_{\substack{S\subseteq N\backslash \{j\}\\ S\neq \emptyset}} \frac{|S|!(n-|S|-1)!}{n!}[ tr\_er(S,m) - tr\_er(S\cup \{j\},m)]  \\
    &=&  \frac{tr\_er(\{j\},m)}{n}  -  \sum\limits_{\substack{S\subseteq N\backslash \{j\}\\ S\neq \emptyset}} \frac{|S|!(n-|S|-1)!}{n!}[ tr\_er(S,m) - tr\_er(S\cup \{j\},m)]  \\
    &=&  \frac{tr\_er(\{j\},m)}{n}  +  \sum\limits_{\substack{S\subseteq N\backslash \{j\}\\ S\neq \emptyset}} \frac{|S|!(n-|S|-1)!}{n!}\underbrace{[tr\_er(S\cup \{j\},m) - tr\_er(S,m)] }_{\leq 0}\\
    &\leq & \frac{tr\_er(\{j\},m)}{n} \leq  tr\_er(\{j\},m).
\end{eqnarray*}
\end{footnotesize}
\qed
\end{proof}

\subsection{Proof of Proposition \ref{prop: 2d_SV_neg}} \label{app: 2d_SV_neg} 
\begin{proof}
From Theorem \ref{thm: appor}, the apportioning of the total training error is:
$$
e_j(m) = \frac{\tilde{c}(m)}{n} - \phi_j(N,v(\cdot,m)) ~~\forall j \in N.
$$
Substituting the value of $\phi_j(N,v(\cdot,m))$ in above equation, we have
\begin{eqnarray*}
    e_j(m) &=&  \frac{tr\_er(\{j\},m)}{n}  +  \sum\limits_{\substack{S\subseteq N\backslash \{j\}\\ S\neq \emptyset}} \frac{|S|!(n-|S|-1)!}{n!}[tr\_er(S\cup \{j\},m) - tr\_er(S,m)].
\end{eqnarray*}
When $n=2$, we have,
\begin{eqnarray*}
 e_1(m) &=& \frac{1}{2}tr\_er(\{1\},m) + \frac{1}{2}(tr\_er(\{1,2\},m)-tr\_er(\{2\},m)) \\
& \leq & \frac{g}{2} +\frac{g^{\prime}-G}{2} \leq g - \frac{G}{2}. 
\end{eqnarray*}
The last inequality along with the condition that $\frac{G}{2}\geq g$ implies that $ e_1(m)\leq 0.$ Similarly, for feature $2$, the apportioning of total training error is given by:
\begin{eqnarray*}
e_2(m) &=& \frac{1}{2}tr\_er(\{2\},m) + \frac{1}{2}(tr\_er(\{1,2\},m)-tr\_er(\{1\},m)) \\
&=& \frac{1}{2}(G+g^{\prime}-g) \geq 0 
\end{eqnarray*} 
The last inequality holds because $G \geq \frac{G}{2}\geq g$.
\qed
\end{proof} 


\subsection{Proof of Theorem \ref{thm: dimen_gen}} \label{app: dimen_gen}
\begin{proof}
Given that $A$ is the set of all feature indices whose class conditional mean is $0$, i.e., $E[X_k|Y=y]=0,~ y\in \{-1,1\},~ k\in A$ and the variance is small, i.e., $var(X_k|Y=y)<\epsilon_k$ for some $\epsilon_k>0$ and for all  $y\in \{-1,1\}$ and $k\in A$. Also, features with indices in $A$ are independent of all other features. 
Now, consider,
\begin{equation}
\begin{aligned}
&\mathbb{P}[(X_j,X_k): X_j \in \mathbb{R}, |X_k| \leq \epsilon_k, j \in A^c, k \in A]\\
&=\mathbb{P}[(X_j,X_k): X_j \in \mathbb{R}, |X_k| \leq \epsilon_k, j \in A^c, k \in A | Y=1]P[Y=1]  \\
& + \mathbb{P}[(X_j,X_k): X_j \in \mathbb{R}, |X_k| \leq \epsilon_k, j \in A^c, k \in A | Y= -1]P[Y=-1]\\
&=\mathbb{P}[X_j : X_j \in \mathbb{R}, j \in A^c | Y=1] \mathbb{P}[X_k: |X_k|\leq \epsilon_k, k \in A | Y=1] \mathbb{P}[Y=1] \\ 
& +\mathbb{P}[X_j : X_j \in \mathbb{R}, j \in A^c | Y=-1] \mathbb{P}[X_k: |X_k|\leq \epsilon_k, k \in A | Y=-1] \mathbb{P}[Y=-1] ~~ \text{($\because$ cond. ind.)} \end{aligned}
\end{equation}

Now, $\mathbb{P}[X_j : X_j \in \mathbb{R}, j \in A^c | Y=y] = 1, \text{for}~~ y=\{-1,1\}$ since support of $\mathbf{X}$ is $\mathbb{R}^n$, therefore

\begin{equation}
\begin{aligned}
&= \mathbb{P}[X_k: |X_k|\leq \epsilon_k, k \in A | Y=1] \mathbb{P}[Y=1] + \mathbb{P}[X_k: |X_k|\leq \epsilon_k, k \in A | Y=-1] \mathbb{P}[Y=-1] \\
&= \left[\prod_{k\in A} \mathbb{P}[|X_k|\leq \epsilon_k| Y=1] \right] \mathbb{P}[Y=1] + \left[\prod_{k\in A} \mathbb{P}[|X_k|\leq \epsilon_k| Y=-1] \right] \mathbb{P}[Y=-1] \\
& = \prod_{k\in A} \mathbb{P}[|X_k|\leq \epsilon_k| Y=y] \left[ \mathbb{P}[Y=1] + \mathbb{P}[Y=-1] \right] \\
&\geq \prod_{k\in A} \left(1- \frac{var(X_k|Y=y)}{\epsilon_k^2} \right).
\end{aligned}
\end{equation}
The last inequality is obtained by using Chebychev's high probability bound.
\qed
\end{proof}

\subsection{Proof of Lemma \ref{lem: eta-normal}} \label{app: eta-normal}
\begin{proof}
Consider the in-class probability $\eta(\mathbf{x})$ given below:
\begin{equation}
\begin{scriptsize}
\begin{aligned}
& \mathbb{P}(Y=1|\mathbf{X}=\mathbf{x})
\\&= \frac{\mathbb{P}(Y=1,\mathbf{X}=\mathbf{x})}{\mathbb{P}(\mathbf{X}=\mathbf{x})}
\\ &=\frac{\mathbb{P}(\mathbf{X}=\mathbf{x}|Y=1)\mathbb{P}(Y=1)}{\sum\limits_{y \in \{-1,1\}}\mathbb{P}(\mathbf{X}=\mathbf{x}|Y=y)\mathbb{P}(Y=y)} \\
&= \frac{\frac{1}{\sqrt{2\pi |\Sigma_+|}}\exp\left[-\frac{1}{2}(\mathbf{x}-\boldsymbol{\mu}_+)^T\Sigma_+^{-1}(\mathbf{x}-\boldsymbol{\mu}_+)\right]  p}{\frac{1}{\sqrt{2\pi |\Sigma_+|}}\exp\left[-\frac{1}{2}(\mathbf{x}-\boldsymbol{\mu}_+)^T\Sigma_+^{-1}(\mathbf{x}-\boldsymbol{\mu}_+)\right] p + \frac{1}{\sqrt{2\pi |\Sigma_-|}}\exp\left[-\frac{1}{2}(\mathbf{x}-\boldsymbol{\mu}_-)^T\Sigma_-^{-1}(\mathbf{x}-\boldsymbol{\mu}_-)\right]  (1-p)} \\
&= \frac{1}{1+ \frac{1-p}{p}\sqrt[]{\frac{|\Sigma_+|}{|\Sigma_-|}}\exp\left[-\frac{1}{2}[(\mathbf{x}-\boldsymbol{\mu}_-)^T\Sigma_-^{-1}(\mathbf{x}-\boldsymbol{\mu}_-) - (\mathbf{x}-\boldsymbol{\mu}_+)^T\Sigma_+^{-1}(\mathbf{x}-\boldsymbol{\mu}_+)] \right]}
\end{aligned}
\end{scriptsize}
\end{equation}
Since in our case $\Sigma_+ = \Sigma_-  =\Sigma =a \mathbf{I} $ 
\begin{equation}
\begin{small}
\begin{aligned}
&=\frac{1}{1+ \frac{1-p}{p}\exp\left[-\frac{1}{2}[(\mathbf{x}-\boldsymbol{\mu}_-)^T\Sigma^{-1}(\mathbf{x}-\boldsymbol{\mu}_-) - (\mathbf{x}-\boldsymbol{\mu}_+)^T\Sigma^{-1}(\mathbf{x}-\boldsymbol{\mu}_+)] \right]}\\
&= \frac{1}{1+ \frac{1-p}{p}\exp \left[-\frac{1}{2}[\mathbf{x}^T(\Sigma^{-1}-\Sigma^{-1})\mathbf{x}- 2 \mathbf{x}^T \Sigma^{-1}\boldsymbol{\mu}_- + 2 \mathbf{x}^T \Sigma^{-1}\boldsymbol{\mu}_+ +  \boldsymbol{\mu}^T_-\Sigma^{-1} \boldsymbol{\mu}_- -\boldsymbol{\mu}_+^T\Sigma^{-1}\boldsymbol{\mu}_+]\right]} \\
& = \frac{1}{1+ \frac{1-p}{p}\exp\left[-\frac{1}{2}[-2 \mathbf{x}^T \Sigma^{-1}(\boldsymbol{\mu}_- - \boldsymbol{\mu}_+) +  \boldsymbol{\mu}^T_-\Sigma^{-1} \boldsymbol{\mu}_- -\boldsymbol{\mu}_+^T\Sigma^{-1}\boldsymbol{\mu}_+]\right]} \\
& = \frac{1}{1+ \frac{1-p}{p}\exp\left[-\frac{1}{2}[\frac{4}{a} \sum\limits_{j \in A_o} x_j {\mu}_{+,j}]\right]}\\
& = \frac{1}{1+ \frac{1-p}{p}\exp\left[\frac{-2}{a} \sum\limits_{j \in A_o} x_j {\mu}_{+,j}\right]}.
\end{aligned}
\end{small}
\end{equation}
The second  last equality follows from the fact that $\Sigma = a\mathbf{I}$ and $\mu_{-,j} =  -\mu_{+,j}$ when $j$ is odd. 
\qed
\end{proof}


\section{More on classification game $(N,v(\cdot,m))$} \label{app: properties_classificationgame}

In this section, we first provide the proof of Proposition \ref{prop: game monotonic} that shows the monotonicity of the characteristic function $v(\cdot,m)$ of classification game.
\subsection{Proof of Proposition \label{proof: game monotonic}}
\begin{proof}
Consider the optimization problem in Section \ref{training_error_fn} solved to obtain $tr\_er(T,m)$ and $tr\_er(S,m)$, say $P_T$ and $P_S$ for coalitions $T$ and $S$ respectively. Now if $ S \subseteq T$, then  in addition to the variables in the optimization problem $P_S$, the optimization problem $P_T$ will have extra variables to solve for. However, a feasible (including optimal) solution in $P_S$ will still remain feasible for $P_T$ by assigning the extra variables a zero value. 
This implies that minimization in $P_T$ is over a larger feasible set and the objective value of $P_T$ (i.e., $tr\_er(T,m)$) would be upper bounded by the objective value of $P_S$ (i.e., $tr\_er(S,m)$). Therefore, the training error due to features in $T$ will be smaller than that of the training error due to the features that come from all its subset, i.e.,
\begin{equation}
\label{c_structure_10}
\forall S \subseteq T \subseteq N, ~~ tr\_er(T,m) \leq tr\_er(S,m).
\end{equation}
The result follows by using $tr\_er(\emptyset,m) =\tilde{c}(m) \geq 0$ and the transformation given in Eq. \eqref{value_to_cost}.
\qed
\end{proof}

Next, we provide definitions of some important classes of cooperative games, viz., superadditive games and convex games. 

\begin{defi}[Superadditive game \cite{Narahari,peleg2007introduction}]
A cooperative game $(N,v)$ is said to be superadditive if  $\forall ~S,T\subseteq N,~ S\cap T = \emptyset, ~~ v(S\cup T) \geq v(S) + v(T)$.
\end{defi}
\begin{defi}[Convex game \cite{Narahari,peleg2007introduction}]
A cooperative game $(N,v)$ is said to be convex if $\forall~ S,T\subseteq N$, $v(S\cup T) + v(S \cap T) \geq v(S) + v(T)$.
\end{defi}

The next natural question after forming a coalition is about the allocation of  the coalition value among the players. This is achieved by a solution concept. Some examples of solution concept are, Shapley value, Nucleolus, Core, etc. As we are focusing on Shapley value and core, we first provide some details about Shapley value axioms and their interpretation in Section \ref{subsec: SV_axioms_Young}. Then, we provide definition of core and how to check whether it is empty or not in Section \ref{subsec: Core_def_BS_char}.

\subsection{Shapley value: An axiomatic approach \cite{young1985monotonic}} \label{subsec: SV_axioms_Young}
In this section, we provide details of Young's axiomatization of Shapley value (\cite{young1985monotonic}) which is based on the following axioms. If $\phi$ denotes the allocation of grand coalition worth, $v(N)$, in the game $(N,v)$ then
\begin{itemize}
\item \textbf{Efficiency:} $\sum\limits_{i\in N} \phi_i(v) = v(N)$
\item \textbf{Symmetry:} If $v(S\cup i) = v(S \cup j) ~~\forall S \subseteq N \setminus \{i,j\}, i,j \in N$, then $\phi_i(v) = \phi_j(v)$
\item \textbf{Marginality:} If two games $(N,v) ~ and ~ (N,w)$ are such that  $v(S\cup i) - v(S) = w(S\cup i) - w(S),~  \forall S \subseteq N \setminus \{i\}$, $i \in N$ then $\phi_i(v) = \phi_i(w)$.
\end{itemize}
\cite{young1985monotonic} proved that the only function that satisfies the above axioms is the Shapley value which is given by:
\begin{equation} \label{eq: Shapley value_appendix}
\phi_j(v) =  \sum\limits_{S\subseteq N\backslash \{j\}} \frac{|S|!(n-|S|-1)!}{n!}[v(S\cup \{j\}) - v(S)], ~ \forall j \in N.
\end{equation}
{\em Interpretation of Young's axioms for Shapley value in classification game: } Consider $ \pi: N \mapsto N$, a permutation of the feature set $N$. Then, the anonymity (symmetry) property of the Shapley value requires that the contribution of a feature $j$ in the total value should be equal to the contribution of the feature $\pi(j)$ in the total value, i.e., $\phi_{j}(N,v(\cdot,m)) = \phi_{\pi(j)}(N,v(\cdot,m))$. In classification, this implies that Shapley value allocation is not dependent on a specific permutation of the features. The efficiency of the Shapley value $\phi_j(N,v(\cdot,m)),~ j \in N$ implies its unique and equitable distribution of the total worth $v(N,m)$ among the players (features) of the game without any deficit or surplus. Marginality property says that if two games with same player set but different value function have equal marginal contribution of a feature $j$, then Shapley value of feature $j$ is equal for the two games. In classification, consider two games based on datasets $D_1$ and $D_2$ sampled from common distribution $\mathcal{D}$. The marginal contribution of a feature $j$ in two games $(N,v_{D_1}(\cdot,m))$ and $(N,v_{D_2}(\cdot,m))$ was observed to be same and hence the marginality property is valid. 

\subsection{Core and its characterization via Bondareva-Shapley theorem} \label{subsec: Core_def_BS_char}

In this section, we describe two important properties, viz., individual rationality, and collective rationality of any allocation, $a$. Towards this, we define the imputation set $I(v)$ as follows:
\begin{equation} \label{eq: imputation_set}
I(v) := \{a\in \mathbb{R}^n: \sum\limits_{j=1}^{n}a_j = v(N), ~~ a_j \geq v(\{j\}) ~ \forall j \in N\}. 
\end{equation}
Next, we define another solution concept, core, that is widely used for allocating the worth among the players. The core, a subset of imputation set, requires that in addition to individual rationality, the allocation should be coalitionally rational.
\begin{defi}[Core \cite{Narahari,peleg2007introduction}] \label{core}
For a cooperative game $(N,v)$ the core $\mathbb{C}(v)$ is defined below:
\begin{equation}\label{core_v}
\begin{footnotesize}
\mathbb{C}(v) := \left\lbrace a \in \mathbb{R}^n: \sum\limits_{j \in N}a_j =v(N), \sum\limits_{j\in S}a_j \geq v(S), ~~ \forall S\subseteq N \right\rbrace.
\end{footnotesize}
\end{equation}
\end{defi}
To check whether the core is empty or not, we use the Bondareva-Shapley theorem \cite{Narahari}. 
\\
\textbf{Bondareva-Shapley characterization:}  Consider the following LP  
\begin{alignat}{3} \nonumber
& \underset{\mathbf{x}}{\min} \sum\limits_{j=1}^{n}x_j \\ \nonumber
 \textrm{s.t. } & \sum\limits_{j\in S} x_j  \geq v(S) ~~ \forall S\subseteq N \\ \label{eq: SV_Bondareva_LP} 
& (x_1, x_2, \ldots, x_n) \in \mathbb{R}^n. 
\end{alignat}
Let $(x_1^*, x_2^*, \ldots, x_n^*)$  be an optimal solution to above LP. If the feasible set of above LP is non-empty then this LP will definitely posses a solution. This is because of the structure of the inequalities, i.e., all inequalities are of the greater than or equal to type. Hence, according to Bondareva-Shapley characterization, if
\begin{enumerate}
    \item \label{sb_cond1} $x_1^* + x_2^* + \ldots + x_n^* =v(N)$, the core is non-empty and all the solutions of above LP will constitute the core.
    \item \label{sb_cond2} $x_1^* + x_2^* + \ldots + x_n^* >v(N)$, the core is empty.
\end{enumerate}
It is well established that core exists for a convex game, and Shapley value lies inside it. But in general, it need not. 

\subsection{Properties of various UCI dataset classification based games}

In this section, we first provide a summary of our investigation on various game theoretic properties of  classification game. Then, we consider some UCI dataset based classification games and check whether these properties hold for them or not.
\begin{rem}\label{remark: super_additive}
The classification game $(N,v(\cdot,m))$ need not be superadditive. We present following counter-example to support our claim.
\begin{example}
Consider the UCI dataset Titanic with three features and 2201 data points. We have $(3,v(\cdot,m))$ as our classification game $(m=2201)$. The characteristic function for this game obtained by solving 7 LPs  and using the transformation given in Eq. (\ref{value_to_cost}) are as follows:
$$v(\{\emptyset\},m) = 0,  v(\{1\},m) = 0 , v(\{2\},m) = 0.0056, v(\{3\},m) = 0.1977, $$ 
$$v(\{1,2\},m) = 0.006, v(\{1,3\},m) = 0.1977, v(\{2,3\},m)= 0.1977, $$
$$ v(\{1,2,3\},m) = 0.1977.$$
If $S= \{2\}$, $T = \{3\}$, then we have 
$ v(S\cup T,m) < v(S,m) + v(T,m).$ This violates the superadditivity condition.
\end{example}
\end{rem} 
\begin{rem}\label{remark: sv_not_in_IP}
For classification game $(N,v(\cdot,m))$, Shapley value $\phi(v(\cdot,m))$ may or may not belong to imputation set. Following counter example supports this claim.
\begin{example}
Again consider the UCI dataset Titanic game $(3,v(\cdot,m))$. The characteristic function values for this game are given in the Remark \ref{remark: super_additive}. The Shapley values for the game $(N,v(\cdot,m))$ are $[0.0; 0.00227; 0.195820082]$.
Clearly, $\phi_2(m) = 0.00227 < v({2},m)= 0.0056$ and $\phi_3(m)=0.19582 < v({3},m) = 0.1977$ implying that the individual rationality condition is violated. Hence, Shapley values of UCI dataset Titanic based game does not belong to imputation set.
\end{example}
\end{rem} 
\begin{rem}
\label{remark: core non empty?}
For the classification game $(N,v(\cdot,m))$, core may or may not be empty. We present two examples: one where the core of classification game is empty and another where the core is non-empty.
\begin{example}[Core is empty]\label{core_empty}
Consider the UCI dataset Pima with eight features and 768 data points. We have $(8,v(\cdot,m))$ as our classification game $(m=768)$. Solving LP given in \eqref{eq: SV_Bondareva_LP}, we obtained [0.010981571, 0.12767811, 0.0018904514, 0.000620087,  0.0015797973, 0.03277906, 0.0093950062,0.00027768529] as an optimal solution, and the condition (\ref{sb_cond2}) of the Bondareva-Shapley characterization is satisfied (because $v(N,m)=0.182679581724$) and hence the core is empty.
\end{example}
\begin{example}[Core is non-empty]\label{core_non_empty}
Consider the UCI dataset Thyroid with five features and 215 data points. We have $(5,v(\cdot,m))$ as our classification game $(m=215)$. Solving LP given in \eqref{eq: SV_Bondareva_LP}, we obtained [0.0038691925, 0.11742864, 0.10407924, 0.2110127, 0.070626081] as an optimal solution, and the condition (\ref{sb_cond1}) of the Bondareva-Shapley characterization is satisfied (because $v(N,m)=0.37435706921$) and hence the core is non-empty.
\end{example}

\end{rem}

\begin{rem} \label{remark: game_notconvex}
Classification game $(N,v(\cdot,m))$ need not be convex. The following UCI dataset based classification game provides an example of a non-convex game with Shapley value belonging to core.
\begin{example}
Consider the UCI dataset Phoneme with five features and 5404 data points. We have $(5,v(\cdot,m))$ as our classification game $(m=5404)$.
Let us consider the following coalitions $S =\{3,4\}$ and $T=\{1,2,4\}$. The values $v(\cdot,m)$ for these coalition after solving the corresponding LP's and taking the transformation are $v(S,m) =0.0297, v(T,m) =0.0361 , v(S\cup T,m)=0.05212, v(S\cap T,m)=0$. It is easy to check that the convexity condition i.e. 
$ v(S \cup T,m) + v(S \cap T,m) \geq v(S,m)+ v(T,m)~~~ \forall S, T \subseteq N $ is violated and hence the game is not convex.
\end{example}
However, in above UCI dataset Phoneme, it is easy to verify that in-spite of a classification game being non-convex, the Shapley value belongs to the core $\mathbb{C}(v(\cdot,m))$.
\end{rem} 
Next, we provide a summary of some properties of classification game $(N,v(\cdot,m))$ corresponding to various UCI datasets in Table \ref{table: Properties of game for diff data sets}. It also shows whether $e_j(m), ~ j\in \{1,\ldots,n\}$ are Coalitionally  Rational (CR), i.e., $\sum\limits_{j\in S}e_j(m) \leq tr\_er(S,m) ~\forall S\subseteq N.$ We also check whether a dataset has negative valued SVEA $e_j(m)$ for some $j\in N.$ The importance of such features is provided in Section \ref{sec: SVEA_implications}.

\begin{table}[htbp!]
\setlength{\tabcolsep}{2pt}
\centering
{\tiny
\begin{tabular}{|l|r|r|r|r|r|r|}
\hline
\textbf{Dataset(n)} & \textbf{$\phi_j(m) \in I(v(\cdot,m))$} & \textbf{$\phi_j(m) \in \mathbb{C}(v(\cdot,m))$} & \textbf{$e_j(m)$ is CR} & \begin{tabular}[c]{@{}c@{}}\textbf{$(N,v(\cdot,m))$} \\ Convex \end{tabular} &  \textbf{$ \mathbb{C}(v(\cdot,m)) \neq \emptyset$} & \textbf{$e_j(m) <0$} \\ \hline
\textbf{Haberman(3)} & $\checkmark$ & $\checkmark$ & $\checkmark$ & $\checkmark$ & $\checkmark$ & $\times$ \\ \hline
\textbf{Titanic(3)} & $\times$ & $\times$ & $\checkmark$ & $\times$ & $\checkmark$ & $\times$ \\ \hline
\textbf{Phoneme(5)} & $\checkmark$ & $\checkmark$ & $\checkmark$ & $\times$ & $\checkmark$ & $\times$ \\ \hline
\textbf{Thyroid(5)} & $\times$ & $\times$ & $\checkmark$ & $\times$ & $\checkmark$ & $\checkmark$ \\ \hline
\textbf{Bupa(6)} & $\checkmark$ & $\checkmark$ & $\checkmark$ & $\times$ & $\checkmark$ & $\times$ \\ \hline
\textbf{Pima(8)} & $\checkmark$ & $\times$ & $\checkmark$ & $\times$ & $\times$ & $\checkmark$ \\ \hline
\textbf{BreastCancer(9)} & $\times$ & $\times$ & $\checkmark$ & $\times$ & $\checkmark$ & $\times$ \\ \hline
\textbf{Magic(10)} & $\times$ & $\times$ & $\checkmark$ & $\times$ & - & $\checkmark$ \\ \hline
\textbf{Heart(13)} & $\times$ & $\times$ & $\times$ & $\times$ & - & $\checkmark$ \\ \hline
\textbf{German(20)} & $\times$ & $\times$ & $\checkmark$ & $\times$ & - & $\times$ \\ \hline
\textbf{Spambase(57)} & $\times$ & $\times$ & $\times$ & $\times$ & - & $\checkmark$ \\ \hline
\textbf{Iris0$^*$(4)} & $\times$ & $\times$ & $\Delta$ & $\times$ & $\checkmark$ & $\checkmark$ \\ \hline
\textbf{Banknote$^*$(4)} & $\times$ & $\times$ & $\checkmark$ & $\times$ & $\checkmark$ & $\checkmark$ \\ \hline
\textbf{Wdbc$^*$(30)} & $\times$ & $\times$ & $\times$ & $\times$ & - & $\checkmark$ \\ \hline
\end{tabular}}
\vspace{0.2 cm}
\caption{Various properties of game for different UCI data sets with $n$ (features). The datasets with $^*$ are almost linearly separable. Also, cells with $-$ are the ones where due to high dimension all $2^n$ training error function $tr\_er(S), \forall S\subseteq N$ cannot be computed and hence Bondareva-Shapley characterization cannot be used for checking the non-emptiness of the Core. It was observed that for Iris0 dataset with $\Delta$, some $e_j$ is not CR due to the fact that $tr\_er(S)= 0~ \forall S\subseteq N\backslash\{\{1\},\{2\}\}$. As we have shown in Section \ref{sec: SVEA_implications}, for some datasets $e_j <0$ for at least one feature $j$ and such features are important for classification.}
\label{table: Properties of game for diff data sets}
\end{table}

\section{Algorithm : ShapleyValue-Aprx} \label{app: approx_algo_SV}
As Shapley value of a feature $j$ is the average marginal contribution to all possible coalitions, computing it is not easy. In general, the problem of computing the Shapley value is known to be NP-hard \cite{faigle1992shapley}. Also, it has high space complexity due to the space requirement of storing $n!$ permutations or $2^n -1$ characteristic function values.

In this section, we present the approximation algorithm for Shapley value by \cite{castro2009polynomial} and how it can be used to obtain the Shapley value of features in the classification game $(N,v(\cdot,m))$.
    
An alternative definition of Shapley value \cite{Narahari} is in terms of all possible orders of the players $N$. Suppose $\pi : \{1,\ldots,n\} \mapsto \{1,\ldots,n\}$ be a permutation and $PermSet(N)$ be the set of all possible permutations with player set $N$. Given a permutation $\pi$, let us denote by $Pred^j(\pi)$  the set of all predecessors of player $j$ in the permutation $\pi$, i.e., $Pred^j(\pi) = \{\pi(1),\ldots,\pi(k-1)\},$ if $j = \pi(k)$. Therefore, the Shapley value can be expressed as follows:
\begin{equation} \label{eq: SV_altformula}
\begin{footnotesize}
\phi_j(m) = \sum\limits_{\pi \in PermSet(N)} \frac{1}{n!}\left[v(Pred^j(\pi)\cup \{j\},m) - v(Pred^j(\pi),m)\right], ~\forall j \in N.
\end{footnotesize}
\end{equation}	
The problem which we face in the exact computation of Shapley value is 2-fold. Firstly, the computation of the $2^n -1$ values of the classification game $(N,v(\cdot,m))$ even for $n$ close to 25 is expensive. Secondly, for summing over the marginal contributions, one needs to either keep track of the powerset (required in Eq. \eqref{eq: Shapley value}) or that of the set of all permutations of features (required in Eq. \eqref{eq: SV_altformula}). For $n>31$ none of the two formulae can be used to compute the set as they reach the maximum data structure limit in implementation languages like Python and R. 
\begin{algorithm}[h!]
\begin{footnotesize}
	\SetAlgoNoLine
	\KwIn{Feature set $N = \{1,2,\ldots,n\}$,  Number of sample permutations $samPerm$, Number of examples $m$,  Set of coalitions $Sam\_co\_set = [()]$. 
	
	\textbf{Initialize:} $v((),m) = 0$, Shapley value estimate $\hat{Sh}_j(m) := 0 ~~ \forall j \in N$.}
	
	Define $tr\_er(\cdot,m)$ on $Sam\_co\_set$ and compute $tr\_er((),m) = \tilde{c}(m)$ using LP in Section \ref{empty_lp}. 
	
    \For {$s=1,2,..,samPerm$}
    {Take $\pi \in PermSet(N)$ with probability $\frac{1}{n!}.$ 
    
     \For {$j=1,2,...,n$}
    {Compute the sets $Pred^j(\pi)$ and $Pred^j(\pi) \cup \{j\},$ 
    
    \If{$Pred^j(\pi)$ \textbf{not in} $Sam\_co\_set$ }
     {Compute $tr\_er(Pred^j(\pi),m)$. \\ 
     Compute $v(Pred^j(\pi),m) = \tilde{c}(m) - tr\_er(Pred^j(\pi),m)$. \\
     Append $Pred^j(\pi)$ to $Sam\_co\_set$. }
    \If{$Pred^j(\pi)\cup \{j\}$ \textbf{not in} $Sam\_co\_set$} 
    {Compute $tr\_er(Pred^j(\pi)\cup \{j\},m).$ \\
    Compute $v(Pred^j(\pi)\cup \{j\},m) = \tilde{c}(m) - tr\_er(Pred^j(\pi)\cup \{j\},m).$ \\
    Append $Pred^j(\pi)\cup \{j\}$ to $Sam\_co\_set.$ }
    $\hat{Sh}_j(m) = \hat{Sh}_j(m)+v(Pred^j(\pi)\cup \{j\},m) - v(Pred^j(\pi),m)$.}}
    $\hat{Sh}_j(m) = \frac{\hat{Sh}_j(m)}{samPerm}, ~~ \forall j \in N$.
\caption{Shapley value approximation scheme} 
\label{alg: SV_appx_algo}
\end{footnotesize}
\end{algorithm}
The approximation algorithm \textbf{ShapleyValue-Aprx}, based on an alternative definition of the Shapley value as in Eq. \eqref{eq: SV_altformula} addresses these issues in following three ways: Firstly, instead of summing over the whole permutation set, we only sum over a sample from the permutation set. Also, the algorithm doesn't need to store the permutation set, i.e., to handle the data structure limits the permutation set is generated uniformly in every round. Secondly, the algorithm does not a priori compute all $2^n -1$ value functions; instead, the value function is calculated for a coalition as and when required in the marginal contribution sum. {Note that, the computation of required $tr\_er(S,m)$ for a coalition $S$ is scalable as it is by an LP.} The estimates $\hat{Sh}_j, ~\forall j \in N$  of Shapley value for the feature set $N$ are shown to be unbiased, consistent and efficient in \cite{castro2009polynomial}. 

The algorithm \textbf{ShapleyValue-Aprx} works as follows: A list $Sam\_co\_set$ of coalitions is initialized with empty coalition. A dictionary $tr\_er(\cdot,m)$ for training error function with key as the elements of $Sam\_co\_set$ is created. Training error when no feature is present i.e. $tr\_er((),m)$ is computed using LP in Section \ref{empty_lp}. Value of empty coalition is then set to zero. In every round, a permutation $\pi$ from set $PermSet(N)$ is picked with probability $\frac{1}{n!}$. For each player $j \in N$, the predecessor set, $Pred^j(\pi)$ is created and checked if that coalition is in $Sam\_co\_set$. If not, then the training error function $tr\_er(Pred^j(\pi),m)$ is computed by solving the corresponding LPs given in Section \ref{training_error_fn}. Values for each coalition using these training errors are computed via the transformation given in Eq. (\ref{value_to_cost}). $Pred^j(\pi)$ is then appended to the list $Sam\_co\_set$. The  above process is repeated for $Pred^j(\pi) \cup \{j\}$ coalition also. The estimate $\hat{Sh}_j(m)$ is updated by adding the marginal contribution of player $j$ for the permutation $\pi$ to the estimate from previous round. After all the rounds are exhausted, we take the average over the number of permutations used, i.e., $samPerm$.

Therefore, algorithm \textbf{ShapleyValue-Aprx} reduces the computational complexity of the Shapley value and also brings down the number of LPs solved.

\section{Details on computational experiments}

\subsection{More details about sample bias robustness technique } \label{app: sample_bias_details}
Here, we provide the details of technique provided in Section \ref{subsec: sample_bias_robust_tech} to make the FSS robust to sample bias.

\begin{enumerate}
    \item Partition the training data into $ss := 30 \times m_s$ subsets where $m_s := 6\times n$ is the sample size of each subset. $m_s$ works as a thumb-rule for training as given in \cite{Beleites2013samplesize}.  
    \item Let the Shapley value and SVEA for $j^{th}$ feature from $r^{th}$ sample subset be denoted by $\phi_{j}^{r}(m_s)$ and $e_{j}^r(m_s)$ for $j \in N$ and $r = 1,\ldots, ss$.
    \item Pick groups of 30 subset of samples without replacement to get $G := \Bigl\lfloor\frac{m}{ss}\Bigr\rfloor$ such groups. For $g^{th}$ group, $g=1,\ldots,G$, for each feature $j \in N$ compute the average of SVEA values across the subsets, i.e., $\bar{e}_j^g = \frac{1}{30}\sum_{r_g}e_{j}^{r_g}(m_s)$ where $r_g$ is the index for sample subsets in $g^{th}$ group. 
    \item Using Central Limit Theorem, we have $\bar{e}_j^{g} \approx Gaussian(\mu_e,\sigma_{e})$ with unknown $\mu_e$ and $\sigma_e$.
    \item Since $\mu_e$ and $\sigma_e$ are unknown, we use $t$-distribution to obtain the $100(1-\alpha)$ confidence interval for $\bar{e}_j^p$ (population mean) using the sample points $\bar{e}_j^{g}, g = 1,\ldots,G$. 
    \item For feature $j \in N$, the interval estimates are $\bar{\bar{e}}_j \pm t^*_{\alpha/2, G-1}\frac{s_j}{\sqrt{G}}$ where $\bar{\bar{e}}_j = \frac{1}{G}\sum_g \bar{e}_j^g$ and $s_j = (\frac{1}{G-1}\sum_g (\bar{e}_j^g - \bar{\bar{e}}_j)^2)^{1/2}$ and $t^*_{\alpha/2, G-1}$ is the upper $\alpha/2$ critical value for the $t$ distribution with $G-1$ degrees of freedom.
\end{enumerate}

\subsection{Additional synthetic data experiments} \label{app: syn_exp_add}
\textbf{1) Synthetic dataset 2 (SD2):} In this example, we consider a scenario where SVEA can be used for FSS even when there is no dimension reduction. The dataset is generated as follows:  generate $3000$ binary class labels $Y$ from Bernoulli distribution ($p=0.5$); draw a $5$-dimensional feature vector $X$ for each label from two different Gaussian distributions: $X|Y=1 \sim N([2; 0.2; 0.3; 1.8; 1],\Sigma)$ $\&$ $X|Y=-1 \sim N([-2; 0.2; 0.3; -1.8; 1], \Sigma)$ where $\Sigma = 10\mathbf{I}_{5\times 5}$. 

The Shapley value based error apportioning $\{e_j(m)\}_{j\in N}$ for 5 features is [-0.0925;  0.1959;  0.1958; -0.0275;$ $  0.1959]. We observe that negative valued SVEA $\{e_j(m)\}_{j\in N}$ can be related to those features whose joint contribution towards classification is significant; individual contribution need not. It can be justified as the test accuracy of only feature one based 1-dimensional linear classifier and only feature four based 1-dimensional linearclassifier is $0.715$ and $0.69$ respectively whereas the test accuracy of feature $\{1,4\}$ based $2$-dimensional linear classifier is $0.818$. It is easy to observe that the accuracies of the linear classifier obtained using the subset of features is comparable to the accuracy of the 5-dimensional classifier obtained via SVM, which is $0.82$. $P_{SV}(\{1,4\}) = 0.997$ confirms that the joint influence of subset $\{1,4\}$ in classification is very high. 
Even though $5$-fold RFECV  and ReliefF algorithm output the same result as above, unlike our SVEA scheme they are based on user-given threshold to decide an optimal number of features; we have a intrinsic data-driven  threshold of $0$ to decide whether the feature is important or not.

Negative value of SVEA for feature 1 and 4 is not due to these features spanning a lower dimensional space because, for this dataset, if one considers feature 2 and feature 3, then using Chebychev's bound $ -2.97 \leq X_2 \leq 3.37$ and $-2.87 \leq X_3 \leq 3.47$ with probability more than $0.99$. Hence, the data is not reducible from $\mathbb{R}^5$ to $\mathbb{R}^2$ even though SVEA values $e_1(m)$ and $e_4(m)$ for feature 1 and 4 are negative.

\textbf{2) Synthetic dataset 3 (SD3):} This example is to demonstrate the ability of SVEA $\{e_j(m)\}_{j \in N}$ to identify the subset of features where the data is predominantly lying. 
We first generate $3000$ binary class labels $Y$ from Bernoulli distribution ($p=0.4$). and then,  a $6$-dimensional feature vector $X$ for each label by drawing a sample such that  $X|Y=1 \sim N([2.5; 0,2.1,0,0,2.6], \Sigma)$ and $X|Y=-1 \sim N([-2.5; 0,-2.1,0,0,-2.6], \Sigma)$. The matrix  $\Sigma$ is symmetric and most of the entries are zero. The only non-zero entries are  $\Sigma_{1,3} = 0.9,  \Sigma_{1,6} = 2.6,  \Sigma_{3,6} = 2$, $\Sigma_{5,5}= 0.002, \Sigma_{k,k} = 5, \text{if}~ k = 1,3,6$ and $\Sigma_{l,l} = 0.001, \text{if} ~l=2,4$.

SVEA $\{e_j(m)\}_{j\in N}$ for 6 features is $[-0.09781946;  0.13057645; -0.02627778; 0.1305889; \\  0.13054749;-0.08510723]$. Now, we  use the result from Theorem \ref{thm: dimen_gen} to show that the probability of the event that feature $k$'s value, $k \in \{2,4,5\}$ is within an $\epsilon_k$-ball, i.e., $P(|X_2| \leq \epsilon_2,|X_4| \leq \epsilon_4,|X_5| \leq \epsilon_5)$ 
is greater than $\prod\limits_{k\in \{2,4,5\}}\left(1- \frac{var(X_k|Y=y)}{\epsilon^2_k}\right)$.
Taking specific value of $\epsilon_k = 0.08, ~k\in \{2,4,5\}$, the joint event $|X_2| \leq 0.08,|X_4| \leq 0.08,|X_5| \leq 0.08$ has probability more than $0.95$. Hence, the dataset is lying in a subspace whose basis corresponds to features $X_j, j \in \{1,3,6\}.$

\textbf{3) Synthetic dataset 4 (SD4):} This example depicts the scenario where SVEA can be used to identify how the inclusion of features non-important for classification (SVEA $e_j >0$) can contribute to the excess $0$-$1$ risk. We first generate a binary class label $Y$ from Bernoulli distribution with parameter $p=0.65$ and then,  a $6$-dimensional feature vector $X$ for the label $Y$ by drawing a sample such that
$X|Y=1 \sim N([2, 0.4 ,2.15, 1,1.1,2.05], \Sigma)$ and $X|Y=-1 \sim N([-2, 0.4,-2.15, 1,1.1,-2.05], \Sigma)$ where $\Sigma$ matrix is same as in Synthetic dataset \textbf{SD(3)}.

As we are interested in providing an interpretation w.r.t unknown data distribution, we consider datasets of varying size from the above data distribution. Also, the datasets are constructed such that the one with a larger number of examples is a superset of the dataset with smaller size. We consider linear classifiers $\mathbf{w}^T\mathbf{x}+b$ and check for the normalized coefficients $w_j, j\in \{1,\ldots,6\}$ to be zero or non-zero. 
Table \ref{Table: Error_appor_Syn_6D} shows that features $1,3,6$ (a variant of set $A_o$ in Section \ref{subsec: eta_error_decomp}) have negative valued SVEA. Using a more general version of Lemma \ref{lem: eta-normal}, we have
\begin{equation} \label{eq: eta_6dexmple3}
\begin{aligned}
\eta(\mathbf{x}) &=\frac{1}{1+\frac{1-p}{p}\exp\left(-\frac{1}{2}(-2\mathbf{x}^T\Sigma^{-1}(\mu_- - \mu_+))\right)} \\
&= \frac{1}{1+0.538\exp{(-0.536x_1 - 0.649x_3 - 2.82x_6)}}.
\end{aligned}
\end{equation}
Hence, the Bayes classifier is only dependent on feature values whose SVEA, $\{e_j(m)\}_{j\in N}$ is negative. Next, we observe the structure of the linear classifiers based on  hinge loss, logistic loss and exponential loss; classifiers (normalized) trained on increasing sample size $m$ are provided in Table \ref{Table: Syn_6d_SVM}, \ref{Table: Syn_6d_LR} and \ref{Table: Syn_6d_ExpERM} respectively. As can be seen, normalized coefficients for feature $2,3$ and $5$, in surrogate loss function based classifiers, are not close to zero. This leads to the conclusion that these classifiers, unlike the Bayes classifier, give weightage to features with $e_j(m)>0$. In Table \ref{Table: Acc_6D}, we present the (test set) accuracy of surrogate loss based classifiers learnt on dataset with increasing sample size $m$ and that of Bayes classifiers, i.e., $(1-\hat{R}_{\text{0-1}}(\hat{f}_l))$ and $(1-\hat{R}_{\text{0-1}}(f_{\text{0-1}}^*))$ respectively. The difference of these test set accuracies gives us $\text{Error }1$ as given in Eq. \eqref{eq: decom_error}. It can be observed in Table \ref{table: error1_sd4} that, even though the magnitude is small, $\text{Error }1$ is positive for almost all surrogate loss based classifiers and all $m$. 
This implies that features with non-negative valued SVEA do contribute to $\text{Error }1$ term.  

As can be seen in Figure \ref{fig: Syn_Real_datasets_interval_estimates},  the interval estimates obtained from sample bias robust technique in Section \ref{subsec: sample_bias_robust_tech} are below origin only for those features which appear in the formula of $\eta(\mathbf{x})$ given in Eq. \eqref{eq: eta_6dexmple3}. This validates our claim about the importance of features whose SVEA is less than 0.


\begin{table}[h!]
\centering
\begin{tabular}{|c|c|c|c|c|c|c|}
\hline
\textbf{m$\backslash$features} & \textbf{1} & \textbf{2} & \textbf{3} & \textbf{4} & \textbf{5} & \textbf{6} \\ \hline
\textbf{500} & 0.01044 & 0.14650 & -0.07845 & 0.14781 & 0.14791 & 0.00378 \\ \hline
\textbf{2000} & -0.01852 & 0.16457 & -0.05623 & 0.16305 & 0.16455 & -0.01415 \\ \hline
\textbf{5000} & -0.02052 & 0.16515 & -0.05062 & 0.16555 & 0.16406 & -0.01236 \\ \hline
\textbf{10000} & -0.01575 & 0.16435 & -0.05446 & 0.16428 & 0.16438 & -0.00997 \\ \hline
\textbf{20000} & -0.01865 & 0.16636 & -0.05647 & 0.16622 & 0.16626 & -0.01253 \\ \hline
\textbf{35000} & -0.01841 & 0.16604 & -0.05444 & 0.16604 & 0.16604 & -0.01344 \\ \hline
\end{tabular}
\vspace{0.2 cm}
\caption{ For Synthetic dataset $(SD4)$ in Section \ref{app: syn_exp_add}, features with $e_j(m)<0$ are the ones which appear in the Bayes classifier via $\eta(\mathbf{x})$ given in Eq. \eqref{eq: eta_6dexmple3}. The above values depict that this phenomenon is prominent even when the sample size $m$ is increased.}
\label{Table: Error_appor_Syn_6D}
\end{table}

\begin{table}[h!]
\centering
\begin{tabular}{|c|c|c|c|c|c|c|}
\hline
\textbf{m$\backslash$coef} & \textbf{$w_1$} & \textbf{$w_2$} & \textbf{$w_3$} & \textbf{$w_4$} & \textbf{$w_5$} & \textbf{$w_6$} \\ \hline
\textbf{500} & 0.32931 & 0.56582 & 0.59305 & -0.14436 & 0.36446 & 0.25695 \\ \hline
\textbf{2000} & 0.13461 & 0.07957 & 0.15461 & 0.95818 & 0.14349 & 0.11379 \\ \hline
\textbf{5000} & 0.17761 & 0.44343 & 0.19392 & 0.69934 & -0.47144 & 0.15132 \\ \hline
\textbf{10000} & 0.19036 & 0.63793 & 0.23270 & 0.65002 & 0.23028 & 0.16466 \\ \hline
\textbf{20000} & 0.18215 & 0.25664 & 0.21781 & 0.45799 & 0.78668 & 0.15775 \\ \hline
\textbf{35000} & 0.41831 & 0.44777 & 0.49562 & 0.03568 & 0.49611 & 0.36259 \\ \hline
\end{tabular}
\vspace{0.2 cm}
\caption{For Synthetic dataset $(SD4)$ in Section \ref{app: syn_exp_add}, Bayes classifier via $\eta(\mathbf{x})$ given in Eq. \eqref{eq: eta_6dexmple3} doesn't depend on features $2,4,5$. However, SVM classifier (with $C=1$), as seen above, has non zero value of $w_j,j=2,4,5$ and hence, contribute to the excess $0$-$1$ risk of SVM classifier.}
\label{Table: Syn_6d_SVM}
\end{table}

\begin{table}[h!]
\centering
\begin{tabular}{|c|c|c|c|c|c|c|}
\hline
\textbf{m$\backslash$coef} & \textbf{$w_1$} & \textbf{$w_2$} & \textbf{$w_3$} & \textbf{$w_4$} & \textbf{$w_5$} & \textbf{$w_6$} \\ \hline
\textbf{500} & 0.48357 & 0.25344 & 0.72844 & -0.07489 & 0.09228 & 0.39646 \\ \hline
\textbf{2000} & 0.23994 & 0.26206 & 0.27568 & 0.82281 & 0.28302 & 0.20158 \\ \hline
\textbf{5000} & 0.28227 & 0.14957 & 0.31771 & 0.82167 & -0.25887 & 0.23421 \\ \hline
\textbf{10000} & 0.24439 & 0.18094 & 0.30115 & 0.76604 & 0.43073 & 0.21092 \\ \hline
\textbf{20000} & 0.21559 & 0.14012 & 0.26106 & 0.55500 & 0.72319 & 0.18630 \\ \hline
\textbf{35000} & 0.38555 & 0.38302 & 0.46065 & 0.37492 & 0.48767 & 0.33773 \\ \hline
\end{tabular}
\vspace{0.2 cm}
\caption{For Synthetic dataset $(SD4)$ in Section \ref{app: syn_exp_add}, Bayes classifier via $\eta(\mathbf{x})$ given in Eq.  \eqref{eq: eta_6dexmple3} does not depend on features $2,4,5$. However, LR classifier (with $C=1$), as seen above, has non zero value of $w_j~j=2,4,5$ and hence, contribute to the excess $0$-$1$ risk of LR classifier.}
\label{Table: Syn_6d_LR}
\end{table}

\begin{table}[h!]
\centering
\begin{tabular}{|c|c|c|c|c|c|c|}
\hline
\textbf{m$\backslash$coef} & \textbf{$w_1$} & \textbf{$w_2$} & \textbf{$w_3$} & \textbf{$w_4$} & \textbf{$w_5$} & \textbf{$w_6$} \\ \hline
\textbf{500} & 0.10013 & 0.57019 & 0.17483 & -0.59722 & 0.51720 & 0.10062 \\ \hline
\textbf{2000} & 0.05134 & 0.32694 & 0.06041 & 0.93642 & 0.08851 & 0.04597 \\ \hline
\textbf{5000} & 0.12060 & 0.10772 & 0.13343 & 0.88682 & -0.39924 & 0.10098 \\ \hline
\textbf{10000} & 0.14923 & 0.13307 & 0.18162 & 0.90821 & 0.29268 & 0.12861 \\ \hline
\textbf{20000} & 0.16415 & 0.25311 & 0.19658 & 0.68720 & 0.61444 & 0.14340 \\ \hline
\textbf{35000} & 0.26931 & 0.21487 & 0.31814 & 0.60514 & 0.59807 & 0.23708 \\ \hline
\end{tabular}
\vspace{0.2 cm}
\caption{For Synthetic dataset $(SD4)$ in Section \ref{app: syn_exp_add}, Bayes classifier via $\eta(\mathbf{x})$ given in Eq.  \eqref{eq: eta_6dexmple3} doesn't depend on features $2,4,5$. However, ExpERM classifier, as seen above, has non zero value of $w_j~j=2,4,5$ and hence, contribute to the excess $0$-$1$ risk of ExpERM classifier.}
\label{Table: Syn_6d_ExpERM}
\end{table}

\setlength{\tabcolsep}{2pt}
\begin{table}[h!]
\centering
\begin{tabular}{|c|c|c|c|c|}
\hline
\textbf{m$\backslash$classifiers} & $\mathbf{1 - \hat{R}_{\text{0-1}}(\hat{f}_{SVM})}$ & $\mathbf{1 - \hat{R}_{\text{0-1}}(\hat{f}_{LR})}$ & $\mathbf{1 - \hat{R}_{\text{0-1}}(\hat{f}_{ExpERM})}$ & $\mathbf{1 - \hat{R}_{\text{0-1}}(f^*_{\text{0-1}})}$ \\ \hline
\textbf{500} & 0.8096 & 0.82 & 0.8144 & 0.8238 \\ \hline
\textbf{2000} & 0.8222 & 0.823 & 0.8222 & 0.8238 \\ \hline
\textbf{5000} & 0.8236 & 0.8226 & 0.8226 & 0.8238 \\ \hline
\textbf{10000} & 0.8238 & 0.8228 & 0.8232 & 0.8238 \\ \hline
\textbf{20000} & 0.8214 & 0.8214 & 0.8206 & 0.8238 \\ \hline
\textbf{35000} & 0.8234 & 0.8226 & 0.823 & 0.8238 \\ \hline
\end{tabular}
\vspace{0.2 cm}
\caption{Test set accuracy for Synthetic dataset $(SD4)$ in Section \ref{app: syn_exp_add}. Each cell contains test set (common and of size 5000 data points) accuracy of a classifier (given in column) trained on sample size $m$ (given in rows). The last column values are for the test set accuracy of the Bayes  classifier $f^*_{\text{0-1}}$ which is independent of $m$. These values can be used to compute $\text{Error }1$ in Eq. \eqref{eq: decom_error}. And if that is positive then, some features non-important for classification identified by Shapley value based error apportioning could contribute to this error. High accuracy of LR and ExpERM classifier relative to that of Bayes classifier can be justified as the accuracies are estimates and not expected values.}
\label{Table: Acc_6D}
\end{table}

\begin{table}[h!]
\centering
\begin{tabular}{|c|c|c|c|}
\hline
\textbf{m/classifier} & \multicolumn{1}{c|}{$\mathbf{Error 1(\hat{f}_{SVM})}$} & \multicolumn{1}{c|}{$\mathbf{Error 1(\hat{f}_{LR})}$} & \multicolumn{1}{c|}{$\mathbf{Error 1(\hat{f}_{ExpERM})}$} \\ \hline
\textbf{500} & 0.0142 & 0.0038 & 0.0094 \\ \hline
\textbf{2000} & 0.0016 & 0.0008 & 0.0016 \\ \hline
\textbf{5000} & 0.0002 & 0.0012 & 0.0012 \\ \hline
\textbf{10000} & 0 & 0.001 & 0.0006 \\ \hline
\textbf{20000} & 0.0024 & 0.0024 & 0.0032 \\ \hline
\textbf{35000} & 0.0004 & 0.0012 & 0.0008 \\ \hline
\end{tabular}
\vspace{0.2 cm}
\caption{This table depicts that using a finite sample linear classifier leads to positive value of Error 1 in the decomposition given in Eq. \eqref{eq: decom_error}. We believe that the non-zero coefficients for feature 2,4 and 5 in Table \ref{Table: Syn_6d_SVM}, \ref{Table: Syn_6d_LR} and \ref{Table: Syn_6d_ExpERM}  contributes to the above error as $\eta$ in Eq. \eqref{eq: eta_6dexmple3} doesn't depend on the above mentioned features.}
\label{table: error1_sd4}
\end{table}

\textbf{4) Synthetic dataset 5 (SD5):} This example illustrates the convergence of Shapley values $\phi_j(m)$ of classification game and the SVEA $e_j(m)$ with increase in sample size $m$. We first generate a binary class label $Y$ from Bernoulli distribution ($p=0.5$) and then,  a $7$-dimensional feature vector $X$ for the label $Y$ by drawing a sample such that  $X|Y=1 \sim N([5, 2.8, 4, 7, 2.8, 3.6,7.5], \Sigma)$ and $X|Y=-1 \sim N([-5,2.8, 3.5, -7, 3.8,3.5,-7.5], \Sigma)$. The matrix  $\Sigma$ is symmetric and most of the entries are zero. The only non-zero entries are  $ \Sigma_{1,3}=\Sigma_{2,5}= \Sigma_{5,7}= 5$, $\Sigma_{1,6} = \Sigma_{2,7}= \Sigma_{3,6}=4$, $\Sigma_{1,7}= 2, \Sigma_{6,7} = 0.1, \Sigma_{3,7} = 8,\Sigma_{4,7} =1$, $\Sigma_{k,k}= 25, \text{if}~ k = 1,2,5,7$ and $\Sigma_{l,l} = 35, \text{if} ~l=3,4,6$.

From  Table \ref{Table: Syn_7d_sv_convergence}, it can be seen that as the sample size increases, the Shapley value of classification game for all features converges. The value function $v(S,m), ~S\subseteq N$ of the classification game is interpreted as the decrease in training error by using features in $S$ for classification relative to classifying without using any feature information. Shapley value of the classification game is then the payoff which a feature gets by participating in the process of classification. Hence, a higher value in long run (with increase in $m$) is an indicator that this particular feature is important.

A more clear interpretation can be given by observing the convergence of SVEA. As the sample size increases, the SVEA $e_j(m)$ converges to the limiting value. Hence, as interpreted in Section \ref{subsec: converg_SV_error}, the limiting values in Table \ref{Table: Apportion_7D} are values of $\{e_j\}_{j\in N}$. Also, as explained in Section \ref{subsec: neg_fss}, the features with negative valued SVEA are the important features for classification.
\begin{table}[!htb]
\centering
\begin{tabular}{|c|c|c|c|c|c|c|c|}
\hline
\textbf{m$\backslash$Feature no.} & \textbf{1} & \textbf{2} & \textbf{3} & \textbf{4} & \textbf{5} & \textbf{6} & \textbf{7} \\ \hline
\textbf{500} & 0.2398 & 0.0009 & 0.0063 & 0.2940 & 0.0093 & 0.0008 & 0.3967 \\ \hline
\textbf{2000} & 0.2425 & 0.0009 & 0.0066 & 0.2930 & 0.0099 & 0.0008 & 0.3902 \\ \hline
\textbf{5000} & 0.2433 & 0.0007 & 0.0049 & 0.2916 & 0.0097 & 0.0007 & 0.3961 \\ \hline
\textbf{10000} & 0.2439 & 0.0006 & 0.0050 & 0.2947 & 0.0109 & 0.0006 & 0.3947 \\ \hline
\textbf{20000} & 0.2470 & 0.0008 & 0.0065 & 0.2988 & 0.0140 & 0.0006 & 0.3943 \\ \hline
\textbf{35000} & 0.2460 & 0.0008 & 0.0066 & 0.2968 & 0.0144 & 0.0006 & 0.3929 \\ \hline
\end{tabular}
\vspace{0.2 cm}
\caption{This table shows the convergence of Shapley value of classification game as the sample size $m$ increases for Synthetic dataset 5 $(SD5)$ in Section \ref{app: syn_exp_add}.  Column heading has feature numbers.}
\label{Table: Syn_7d_sv_convergence}
\end{table}

\begin{table}[h!]
\centering
\begin{tabular}{|c|c|c|c|c|c|c|c|}
\hline
\textbf{m$\backslash$Feature no.} & \textbf{1} & \textbf{2} & \textbf{3} & \textbf{4} & \textbf{5} & \textbf{6} & \textbf{7} \\ \hline
\textbf{500} & -0.1004 & 0.1385 & 0.1331 & -0.1545 & 0.1301 & 0.1386 & -0.2573 \\ \hline
\textbf{2000} & -0.1029 & 0.1387 & 0.1329 & -0.1535 & 0.1296 & 0.1387 & -0.2506 \\ \hline
\textbf{5000} & -0.1034 & 0.1391 & 0.1350 & -0.1517 & 0.1302 & 0.1391 & -0.2563 \\ \hline
\textbf{10000} & -0.1037 & 0.1396 & 0.1352 & -0.1545 & 0.1293 & 0.1396 & -0.2545 \\ \hline
\textbf{20000} & -0.1049 & 0.1413 & 0.1356 & -0.1567 & 0.1280 & 0.1415 & -0.2522 \\ \hline
\textbf{35000} & -0.1042 & 0.1410 & 0.1352 & -0.1550 & 0.1275 & 0.1413 & -0.2510 \\ \hline
\end{tabular}
\vspace{0.2 cm}
\caption{This table shows the convergence of Shapley value based error apportioning $\{e_j\}_{j\in N}$ as the sample size $m$ increases for Synthetic dataset 5 $(SD5)$ in Section \ref{app: syn_exp_add}. This example demonstrates the existence of true unknown hinge risk for each feature, $\{e_j\}_{j\in N}$ as defined in Section \ref{subsec: converg_SV_error}. Column heading has feature numbers.}
\label{Table: Apportion_7D} 
\end{table}

\subsection{Effect of regularization used in $tr\_er(\cdot,m)$ on SVEA} \label{app: linear_with_WO_reg_details}
In this subsection, we provide details of the experiments which we used to claim that with linear classifiers regularization is not helpful for feature subset selection task. Consider the regularized version of $tr\_er(S,m)$ as defined in Section \ref{training_error_fn} with trade-off parameter $C>0$ as follows:
\begin{equation} \label{eq: trer_reg_linear}
\begin{aligned}
&tr\_er(S,m) = \min\limits_{w_{j_1},\ldots,w_{j_r},b_r,\{\xi_i\}_{i=1}^m} C\sum\limits_{i=1}^{m}\xi_i + \frac{1}{2}\Vert \mathbf{w}\Vert^2 \\
& \text{s.t. } y_i \left(\sum\limits_{j \in S}w_{j}x_{ij} + b_{r}\right)  \geq 1 - \xi_i ~~ \forall i = 1,\ldots,m \\
& \xi_i \geq 0 ~~ \forall i = 1,\ldots,m.
\end{aligned}
\end{equation}

Clearly, using Eq. \eqref{eq: trer_reg_linear} to compute $v(S) = tr\_er(\emptyset,m) - tr\_er(S,m)$ with $tr\_er(\emptyset,m)$ as in Section \ref{training_error_fn} can lead to $v(S,m)$ being negative. To avoid this issue, we define $v_{reg}(S,m) = tr\_er(\emptyset,m)- \frac{1}{m}\sum\limits_{i=1}^{m}\xi^*_i$ where $\xi_i^*, i=1,\cdots,m$ is optimal solution of problem in Eq. \eqref{eq: trer_reg_linear}. Even though $v_{reg}(S,m)$ is not shown to be theoretically positive, we observed it to be positive in all our experiments. 

We computed SVEA using $v_{reg}(S,m)$ for various real and synthetic datasets across 5 trials. We tuned the parameter $C$ in the set $\{0.1,1,50,500\}$ for the optimization problem in EQ. \ref{eq: trer_reg_linear} when $S=N$ and used the best value of $C$ obtained $S=N$ in the optimization problem for all other subsets $S\neq N$. We observed that the important feature subset corresponding to those features that have SVEA $e_j(m)<0$ is same irrespective of the fact whether regularization is used or not in the characteristic function. This is verified across 5 trials on the datasets for which Shapley value can be computed exactly. Details available in Table \ref{tab: reg_unreg_compar_exper_lin}. For datasets where algorithm \ref{alg: SV_appx_algo} is used, we observe that the subset $SVEA_{neg}$ varies across trials and is different with and without regularization. We repeated this experiment many times and observed different elements in $SVEA_{neg}$. This phenomenon is possibly not the effect of regularization but that of permutation sampling used while computing Shapley value estimates. Hence, based on our computational experiments, we conclude that, in case of linear classifiers,  regularization in SVEA scheme is not helpful for feature subset selection.

\begin{table}[h!]
\begin{tabular}{|c|c|c|}
\hline
\textbf{Dataset (m,n)} & \textbf{$SVEA_{neg}$ (without reg)} & \textbf{$SVEA_{neg}$ (with reg)} \\ \hline
\textbf{SD2 (3000,5)} & $\{1,4\}$ & $\{1,4\}$ \\ \hline
\textbf{SD3 (3000,6) } & $\{1,3,6\}$ & $\{1,3,6\}$ \\ \hline
\textbf{SD4 (9000,6)} & $\{1,3,6\}$ & $\{1,3,6\}$ \\ \hline
\textbf{Thyroid (215,5)} & $\{4\}$ & $\{4\}$ \\ \hline
\textbf{Pima (768,8)} & $\{2\}$ & $\{2\}$ \\ \hline
\textbf{Heart (270,13)} & \begin{tabular}[c]{@{}c@{}}$\{9,12,13\}; \{3,11,12,13\}; \{3,9,12,13\};$ \\ $\{3,12,13\}; \{3,9,12,13\}$\end{tabular} & \begin{tabular}[c]{@{}c@{}}$\{9,11,12,13\}; \{3,12,13\};\{3,9,12,13\};$ \\ $\{3,12,13\}; \{3,12,13\}$\end{tabular} \\ \hline
\end{tabular}
\caption{Comparison of important feature subset $SVEA_{neg}$  when the characteristic function was defined with and without regularization over 5 different trials (train-test partitioning). For $n<10$, Shapley value is computed exactly and $SVEA_{neg}$ is same. For datasets with $n\geq 10$, use of Shapley value estimates led to difference in the sets obtained with and without regularization.}
\label{tab: reg_unreg_compar_exper_lin}
\end{table}

\subsection{Behaviour and interpretation of kernel (non-linear classifiers in $tr\_er$) based SVEA values} \label{app: kernel_with_reg_details}
In this subsection, we consider non-linear classifiers by using kernel. Formally, let $\phi: \mathbb{R}^n \mapsto \mathbb{R}^z$ with $z>>n$ be the feature map that lifts a given feature vector to a higher dimensional feature space. Then, the regularized $tr\_er(S,m)$ function with feature map $\phi(\mathbf{x})$ is defined as follows: 

\begin{equation} \label{eq: trer_reg_kernel}
\begin{aligned}
&tr\_er_{k,reg}(S,m) = \min\limits_{w_{j_1},\ldots,w_{j_r},b_r,\{\xi_i\}_{i=1}^m} C\sum\limits_{i=1}^{m}\xi_i + \frac{1}{2}\Vert \mathbf{w}\Vert^2 \\
& \text{s.t. } y_i \left(\mathbf{w}^T\phi(\mathbf{x}) + b_{r}\right)  \geq 1 - \xi_i ~~ \forall i = 1,\ldots,m \\
& \xi_i \geq 0 ~~ \forall i = 1,\ldots,m.
\end{aligned}
\end{equation}

Note that regularization in Eq. \eqref{eq: trer_reg_kernel} is necessary to get the feature map dot product term (to be replaced by kernel $\kappa(\mathbf{x},\mathbf{x}\prime)= \phi(\mathbf{x})^T\cdot\phi(\mathbf{x}\prime)$ ) in the dual. Now, $v_{k, reg}(S,m) := tr\_er(\emptyset,m) - tr\_er_{k,reg}(S,m)$ need not be positive. Further, the trick of redefining $v_{k,reg}(S,m)$ using the optimal slack variables $\xi_i, i=1,\cdots,m$ cannot be used  here as the dual solution doesn't provide a closed form expression for the optimal $\xi_i$ values. Hence, we continued using $v_{k,reg}(S,m)$ as defined earlier as an exploratory study.

We performed experiments on some UCI datasets for which the results are summarized in Table \ref{tab: reg_unreg_compar_exper_kernel}. We used radial basis function (Gaussian kernel) defined as $\kappa(\mathbf{x},\mathbf{x}\prime) = \exp{(-\gamma\Vert \mathbf{x} - \mathbf{x}\prime\Vert^2)}$ where $\gamma>0$ is the scale parameter to be tuned using the data. We used two methods to tune the value of $\gamma$: cross validate over a given set $\Gamma = \{0.01,0.1,1,10\}$ or use the most suggested value $\gamma = \frac{1}{n*Var(\mathbf{X})}$ where $Var(\mathbf{X})$ is the variance of the training data point-feature matrix $\mathbf{X}$ to be used. We present the results based on later method as it led to better test accuracies and had consistency in results across trials. The value of C is tuned similarly as in the linear classifier case. From Table \ref{tab: reg_unreg_compar_exper_kernel}, one can observe the variation in identifying the important subset $SVEA_{neg}$ across the trials for Magic and Heart dataset. Also, in Thyroid dataset using rbf kernel leads to $SVEA_{neg} = \{2\}$ which is completely different from the one obtained in linear case ($SVEA_{neg} = \{4\}$). An important point to note for thyroid dataset here is that, even though use of kernels is leading to more than 90\% test accuracy of full dimensional feature set based classifier and important feature subset based classifier, power of classification in the last column is same (0.94).


This erratic behaviour of the SVEA scheme when the training error function is kernelized and regularized can be attributed to the non-monotonic nature of the characteristic function $v_{k,reg}(S,m)$ (arising due to regularization). This monotonicity is important as it implies that the underlying assumption of grand coalition formation $N$ for Shapley value is satisfied. Without monotonicity in the characteristic function, its not easy to justify the use of Shapley value. This problem exists in the case of linear regularized based SVEA but the positivity of $v_{reg}(S,m)$ is a saving grace and the solutions are sensible and understandable. In kernelized and regularized case, the problem is prominent due to $v_{k,reg}(S,m)$ being negative. 

Based on the computations, we would like to suggest that using only linear classifiers in the training error function in Section \ref{training_error_fn}
is good enough to identify the important feature subset unless the data is highly inseparable and there is a domain requirement of using non-linear classifiers. For such cases, one has to resort to redefining the characteristic function $v(S,m)$ to make sure that the monotonicity condition is satisfied.

\begin{table}[h!]
\centering
\setlength{\tabcolsep}{1.5pt}
\renewcommand{\arraystretch}{0.9}
{\scriptsize
\begin{tabular}{|c|c|c|c|c|c|}
\hline
\textbf{Dataset (m)} & $\mathbf{n}$ & \multicolumn{1}{c|}{\textbf{\begin{tabular}[c]{@{}c@{}}Avg Acc \\($\pm$std dev) SVM \end{tabular}}} & \textbf{$SVEA_{neg}$} & \textbf{\begin{tabular}[c]{@{}c@{}}Avg Acc \\ ($\pm$std dev) SVM \\ with $SVEA_{neg}$\end{tabular}} & \textbf{\begin{tabular}[c]{@{}c@{}}$P_{SV}$\\ $(SVEA_{neg})$\end{tabular}} \\ \hline
\textbf{Thyroid (215)} &5 &  0.96 $\pm$ 0.0186 & $\{2\}$ & 0.91 $\pm$ 0.0348 & 0.94 \\ \hline
\textbf{\begin{tabular}[c]{@{}c@{}}Pima Diabetes(768)\end{tabular}} & 8 &  0.76 $\pm$ 0.0126  & \{2\} & 0.74 $\pm$ 0.0214  & 0.97\\ \hline
\textbf{\begin{tabular}[c]{@{}c@{}}Magic(19020)\end{tabular}} & 10 & 0.84 $\pm$ 0.0007 & \{9\};\{1,2,9\};\{9\};\{9\};\{9\} & 0.75 $\pm$ 0.0425
&  0.89\\ \hline
\textbf{Heart (270)} & 13  & 0.81 $\pm$ 0.014  & \begin{tabular}[c]{@{}c@{}} \{3,12,13\};\{3,13\};\\ \{3,12,13\};  \{12,13\}; \\ \{3,12,13\}  \end{tabular} & 0.80 $\pm$ 0.0746 & 0.98\\ \hline
\end{tabular}}
\vspace{0.2 cm}
\caption{Accuracies of the datasets having negative SVEA for features in $SVEA_{neg}$ computed using $v_{k,reg}(S,m)$. The second last column has the accuracy of the SVM classifier (using rbf kernel with $\gamma = \frac{1}{n*Var(X)}$) learnt  only on features in $SVEA_{neg}$. $P_{SV}(SVEA_{neg})$ is the ratio  of accuracies in column 3 and column 5. SVM parameter $C \in \{0.1,1,50,500\}$.}
\label{tab: reg_unreg_compar_exper_kernel}
\end{table}

\subsection{Comparison to $l_1$-regularized squared hinge based ERM, RFECV and ReliefF} \label{app: LASSO_rfecv_reliefF_details}

Before providing the details of the comparison with other feature selection techniques, we show that for a real dataset (Pima Diabetes), negative valued SVEA $e_j(m)$ identifies the features whose joint contribution towards classification is large and not the ones which provide the basis for the lower dimensional subspace. Pima dataset has $e_2(m) <0$ i.e, its $SVEA_{neg} = \{2\}$ with $P_{SV}(SVEA_{neg}) = 0.98$. We computed the class-wise mean (row 1 and row 3 of Table \ref{Table: PIMA_mean_sd}) for each feature in Pima dataset and observed that it is significantly different from 0, and the corresponding standard deviation (row 2 and row 4 of Table \ref{Table: PIMA_mean_sd}) is also not trivial. It indicates that the above phenomena of only 1-feature having the majority of decisive power in classification is not a manifestation of dimension reduction, and feature 2 is an important feature. Domain knowledge also confirms this as feature 2 is the Blood glucose level, which is an almost sufficient test for deciding whether a person has diabetes or not.
\begin{table}[h!]
\centering
\begin{tabular}{|c|c|c|c|c|c|c|c|c|}
\hline
\textbf{Feature no.} & \textbf{1} & \textbf{2} & \textbf{3} & \textbf{4} & \textbf{5} & \textbf{6} & \textbf{7} & \textbf{8} \\ \hline
\textbf{Mean (positive class)} & 4.83 & 140.47 & 70.56 & 22.49 & 97.10 & 34.89 & 0.55 & 36.83 \\ \hline
\textbf{Std dev (positive class)} & 3.71 & 30.37 & 21.41 & 17.79 & 135.85 & 7.43 & 0.37 & 10.87 \\ \hline
\textbf{Mean (negative class)} & 3.27 & 109.22 & 67.87 & 19.62 & 69.06 & 30.21 & 0.43 & 31.13 \\ \hline
\textbf{Std dev (negative class)} & 2.98 & 26.49 & 17.72 & 14.89 & 102.85 & 7.82 & 0.31 & 11.61 \\ \hline
\end{tabular}
\vspace{0.2 cm}
\caption{ Dataset statistics for Pima dataset (768 examples and 8 features). The class-wise mean (row 1 and 3) for all features is different from 0 and there is significant standard deviation (row 2 and 4) implying the data is not residing in lower dimension.  {However, selecting feature 2 that has negative value of SVEA $e_2(m)$ yields a 1-d classifier with comparable accuracy and $P_{SV}(\{2\})$ close to $1$; see Table \ref{Table:real_dataset_SV_neg}}.}
\label{Table: PIMA_mean_sd}
\end{table}

The classical $l_1$ regularization is known to impart sparsity to a classifier and hence, identifies important features. We first show that on the datasets in which SVEA identifies important features given by $SVEA_{neg}$, implementation of $l_1$ regularized squared hinge loss based ERM leads to a linear classifier that doesn't have zero coefficient for any of the features and hence is not able to identify important features. The above conclusion is based on the observations on UCI datsets Thyroid, Pima, and Heart given in Table \ref{table: l1_regu_real}. The table presents results from one trial. We had repeated the experiments 5 times and observed the same phenomenon. Value of the parameter $C$ is chosen from set $\{0.1,1,50,500\}$.

\begin{table}[h!]
\centering
{\footnotesize  
\begin{tabular}{|c|c|c|}
\hline
\textbf{Datasets} & $SVEA_{neg}$ & Coefficients of a linear classifier $\mathbf{\{w_j : j \in N\}}$,  $\mathbf{b}$ \\ \hline
\textbf{Thyroid} & $\{ 4\}$ & $\{-0.129717,  0.275245,   0.322175,  1.433379,  0.213262\}$,  $-0.249558$ \\ \hline
\textbf{Pima} & $\{ 2\}$ &\begin{tabular}[c]{@{}c@{}}$\{0.044308,  0.012393, -0.005553,  0.001036, -0.000439, $  \\ 0.033215, $0.236055,  0.005251\}$, $ -2.946975$\end{tabular} \\ \hline
\textbf{Heart} & $\{ 3,12,13\}$& \begin{tabular}[c]{@{}c@{}}$\{0.011814,  -0.292897,  -0.126404,  -0.004811,  -0.001760 ,$  \\   $0.226762,  -0.099385  0.009803,  -0.312632,  -0.015273,$  \\  $-0.198566,  -0.410042,    -0.151856\}$,  $1.24423077$ \end{tabular} \\ \hline
\end{tabular}}
\vspace{0.2 cm}
\caption{ \footnotesize The above table shows that the coefficients of a linear classifier learnt from an $l_1$-regularized squared hinge loss based ERM are non-zero for all the features. This implies that there is no sparsity due to $l_1$ regularization and we cannot comment on important features. $SVEA_{neg}$ is the set of features with negative valued SVEA. This is the set of features important for classification as identified by SVEA scheme. Hence, in above datasets, $l_1$ regularization based method for feature selection doesn't identify important features, but our SVEA is able to identify the important features given in set $SVEA_{neg}$. This is verified by high value of $P_{SV}$ in Table \ref{Table:real_dataset_SV_neg}. }
\label{table: l1_regu_real}
\end{table}

We also compare these results to an existing FSS technique called Recursive feature elimination with cross validation (RFECV). The results are presented in Table \ref{table: feat_RFECV}. For the datasets in which SVEA identifies important features, i.e., $SVEA_{neg}$ is non- empty, the set of important features from column 2 and column 3 of Table \ref{table: feat_RFECV} have some common features. And for the datasets, where SVEA is indicating that there are no negative $e_j(m)$ but RFECV is selecting only few features as important, we provide an explanation in the ``Comments'' column of Table \ref{table: feat_RFECV}. Comment ``Same accuracy for all features'' means that RFECV arbitrarily picked any one feature as important because RFECV's accuracy is same irrespective of the fact that whether one uses 1 or 2 or 3 features.

\begin{figure*}[!h]
    \centering
    \begin{subfigure}[b]{0.475\textwidth}
        \centering
        \includegraphics[width=1.04\textwidth]{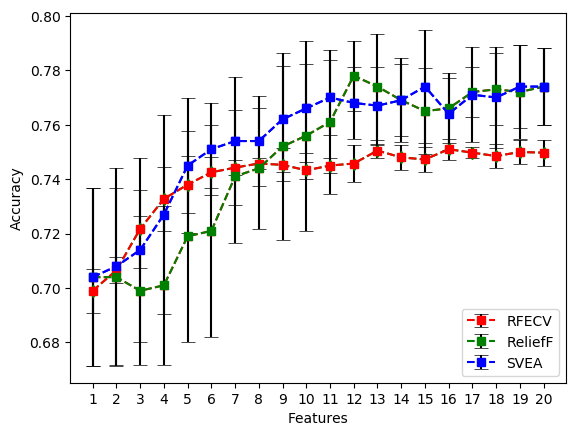}
    \caption*{\footnotesize{Dataset: German (1000,20)}}
        {}    
    \end{subfigure}
    \hfill
    \begin{subfigure}[b]{0.475\textwidth}  
        \centering 
        \includegraphics[width=1.04\textwidth]{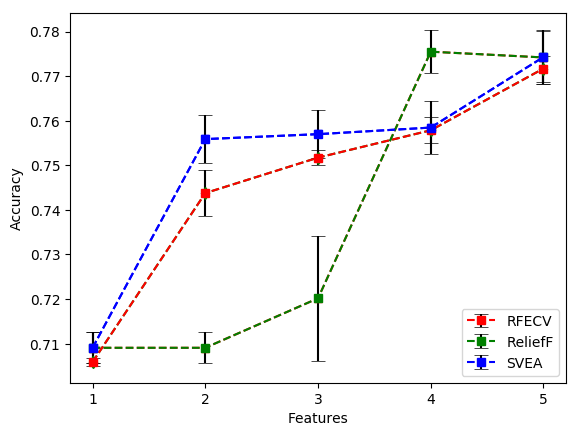}
    \caption*{\footnotesize{Dataset: Phoneme (5404, 5)}}
        {}       
    \end{subfigure}
    \vskip\baselineskip
        \begin{subfigure}[b]{0.475\textwidth}
        \centering
        \includegraphics[width=1.04\textwidth]{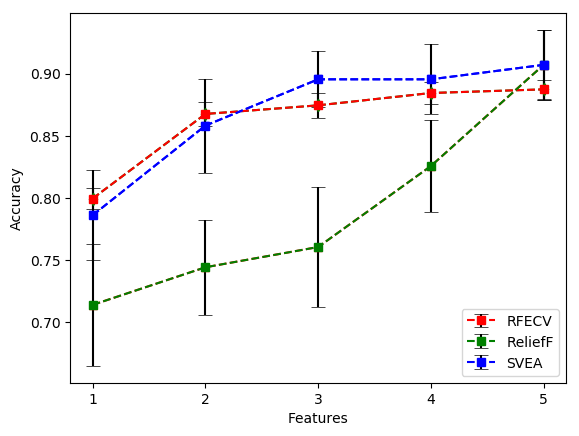}
    \caption*{\footnotesize{Dataset: Thyroid (215, 5)}}
        {}    
    \end{subfigure}
    \hfill
    \begin{subfigure}[b]{0.475\textwidth}   
        \centering 
        \includegraphics[width=1.04\textwidth]{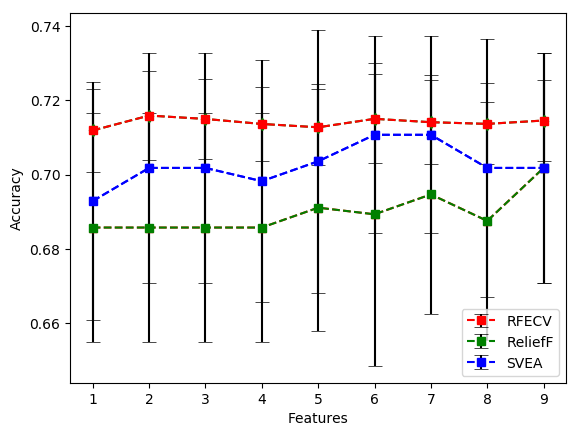}
    \caption*{\footnotesize{Dataset: Breastcancer (277,9)}}
        {}       
    \end{subfigure}
    \vskip\baselineskip
        \begin{subfigure}[b]{0.475\textwidth}
        \centering
        \includegraphics[width=1.04\textwidth]{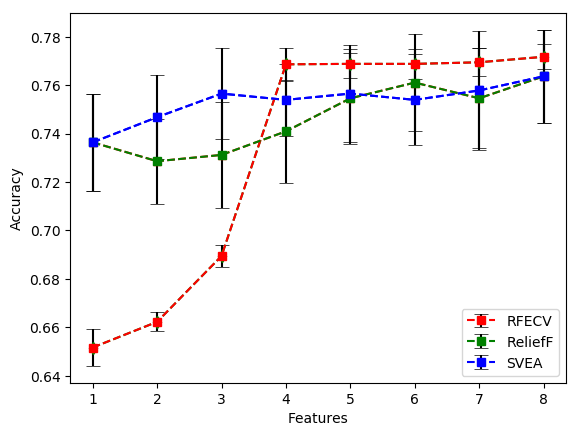}
    \caption*{\footnotesize{Dataset: Pima (768,8)}}
        {}    
    \end{subfigure}
    \hfill
    \begin{subfigure}[b]{0.475\textwidth}   
        \centering 
        \includegraphics[width=1.04\textwidth]{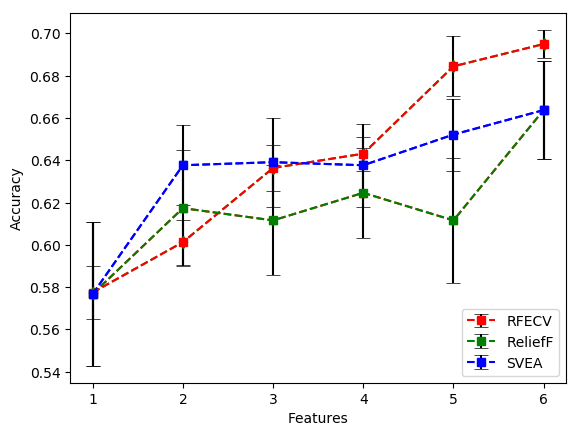}
    \caption*{\footnotesize{Dataset: Bupa (345, 6)}}
        {}       
    \end{subfigure}
    \caption{\footnotesize Plot of test accuracy vs number of features used to train the linear classifier using SVM. For each scheme, we have $95\%$ error bar computed over 5 iterations. More details about the observed behaviour are provided in the comments column of Table \ref{table: feat_RFECV}.} 
    \label{fig: comparison_real_data_add1}
\end{figure*}

Figure \ref{fig: comparison_real_data_add1} provide the comparison of SVEA scheme to RFECV and ReliefF for UCI datasets. In datasets German and Thyroid, clearly SVEA performs better in terms of accuracy. In Breastcancer dataset, RFECV has equal accuracy whether you use only one feature or all 9 features; whereas SVEA is able to identify a feature set of size 6 or 7 that leads to good accuracy.

\setlength{\tabcolsep}{2pt}
\begin{table}[h!]
\centering
{\scriptsize
\begin{tabular}{|c|c|c|c|}
\hline
\textbf{Dataset(n)}& \textbf{$SVEA_{neg}$} & \begin{tabular}[c]{@{}c@{}}RFECV: Important\\  Feature number\end{tabular} & \textbf{Comments} \\ \hline
\textbf{Haberman(3)} & $\emptyset$ & 3 & Same accuracy for all features. \\ \hline
\textbf{Titanic(3)} & $\emptyset$ & 3 & Same accuracy for all features. \\ \hline
\textbf{Phoneme(5)} & $\emptyset$ & All & \begin{tabular}[c]{@{}c@{}}Highest RFECV accuracy was obtained \\ with all 5 feaures.\end{tabular} \\ \hline
\textbf{Thyroid(5)} & $\{4\}$ & 2,3,4 & \begin{tabular}[c]{@{}c@{}}RFECV accuracy with 2 features \\ and 3 features is equal.\end{tabular} \\ \hline
\textbf{Bupa(6)} & $\emptyset$ & All & \begin{tabular}[c]{@{}c@{}}Highest RFECV accuracy was obtained \\ with all 6 feaures.\end{tabular} \\ \hline
\textbf{Pima(8)} & $\{2\}$ & All except 4,5,8 & \begin{tabular}[c]{@{}c@{}}SVM test accuracy with RFECV selected \\ features is 0.771 and with features \\ in $SVEA_{neg}$ is 0.766.\end{tabular} \\ \hline
\textbf{BreastCancer(9)} & $\emptyset$ & 4 & Same accuracy for all features. \\ \hline
\textbf{Heart(13)} & $\{3,12,13\}$ & All except 4,5,8 & \begin{tabular}[c]{@{}c@{}}SVM test accuracy with RFECV selected\\ features and features in $SVEA_{neg}$ is same, \\ i.e., 0.8148.\end{tabular} \\ \hline
\textbf{German(20)} & $\emptyset$ & 7 out of 20 & \begin{tabular}[c]{@{}c@{}}The RFECV accuracy difference between \\ using 7 features and 20 features \\ is less that $0.5\%$.\end{tabular} \\ \hline
\textbf{Spambase(57)} & $a$ & 55 out of 57 & \begin{tabular}[c]{@{}c@{}}After 20 features onwards RFECV accuracy \\ is almost constant.\end{tabular} \\ \hline
\textbf{Wdbc(30)} & $b$ & 12 out of 30 & \begin{tabular}[c]{@{}c@{}}SVM test accuracy with features \\ in $SVEA_{neg}$ and RFECV selected features \\ is same, i.e., 0.9385.\end{tabular} \\ \hline
\textbf{Banknote(4)} & $\emptyset$ & All & \begin{tabular}[c]{@{}c@{}}Highest RFECV accuracy was obtained \\ with all 4 feaures.\end{tabular}  \\ \hline
\textbf{Iris0(4)} & $\{3,4\}$ & 3 & Same accuracy for all features. \\ \hline
\textbf{Magic(10)} & $\{9\}$ & All except 6 and 8 & \begin{tabular}[c]{@{}c@{}}SVEA achieves SVM test accuracy of 0.75\\ by only training on feature number 9 \\ whereas RFECV achieves this accuracy with \\ 4 features.\end{tabular} \\ \hline
\end{tabular}}
\vspace{0.2 cm}
\caption{ \footnotesize This table compares the SVEA scheme to an existing feature selection technique called RFECV using SVM as the estimator and 5 folds cross validation. It was observed that for some datasets taking 2 folds leads to different features as important. The SVEA based FSS scheme is independent of such user given parameters. Also, $\emptyset$ in column 2 means that there is no one dominating feature and hence {\em all} features are important. $a:$ Cardinality of $SVEA_{neg}$ for Spambase (computed using \textbf{ShapleyValue-Aprx}) is less than the set of RFECV based optimal features but the RFECV accuracy (using 2 folds) was same as that obtained using $SVEA_{neg}$ features. $b:$ Cardinality of set $SVEA_{neg}$ for WDBC (computed using \textbf{ShapleyValue-Aprx}) is 13 and it has 10 features common to the set of optimal features provided by RFECV. Since, Iris0 dataset is linearly separable, i.e., $tr\_er(N)=0$, some feature will always have negative valued SVEA to respect the collective rationality property of SVEA. This implies that one cannot interpret $SVEA_{neg} = \{3,4 \}$ here as the set of important features. The value of  $C$ in SVM is chosen from set $\{0.1,1,50,500\}$. Here, $n$ denotes the number of features.}
\label{table: feat_RFECV}
\end{table}

\end{document}